\documentclass[runningheads]{llncs}

\usepackage{eccv}

\usepackage{eccvabbrv}

\usepackage{graphicx}
\usepackage{booktabs}
\usepackage{multirow}
\usepackage{xcolor}
\usepackage{colortbl}
\usepackage{subcaption}
\usepackage[accsupp]{axessibility}  % Improves PDF readability for those with disabilities.

\usepackage{hyperref}

\usepackage{orcidlink}

\begin{document}

\title{VICAL: Vicinal Consistency Alignment for Long-Tailed Visual Recognition} 

\titlerunning{VICAL}

\author{Jianggang Zhu\inst{1,2}\orcidlink{0009-0003-4541-010X} \and
Zheng Wang\inst{3}\orcidlink{0000-0002-6753-6569} \and
Bin Zhu\inst{4}\orcidlink{0000-0002-9213-2611} \and
Yi-Ping Phoebe Chen\inst{5}\orcidlink{0000-0002-4122-3767} \and
Jingjing Chen\inst{1,2}\orcidlink{0000-0003-3148-264X}\thanks{Corresponding author. Email: \{jgzhu20, chenjingjing\}@fudan.edu.cn}
}

\authorrunning{Zhu et al.}

\institute{Institute of Trustworthy Embodied AI, Fudan University, Shanghai, China \and
Shanghai Key Laboratory of Multimodal Embodied AI, Shanghai, China
\and Zhejiang University of Technology, Hangzhou, China
\and Singapore Management University, Singapore, Singapore \and  La Trobe University, Melbourne, Australia
}

\maketitle

\begin{abstract}
Multi-expert models have become the dominant paradigm for long-tailed learning, largely attributed to their presumed ability to benefit from expert diversity. However, we revisit this central assumption and reveal that diversity induced by logit adjustment or explicit regularizers does not guarantee better ensemble accuracy. 
Our work suggests that multi-expert models benefit more from variance reduction than diversity maximization.
We introduce \textbf{VICAL}, a \textbf{VI}cinal \textbf{C}onsistency \textbf{AL}ignment framework that improves long-tailed recognition not by enforcing expert diversity, but by reducing prediction variance.
Specifically, our approach comprises two key components: Self-Consistency Learning and Deep Ensemble Distillation. 
Self-Consistency Learning discourages reliance on unstable high-frequency information, smoothing the local loss landscape and mitigating overfitting, especially for tail classes.
Deep Ensemble Distillation promotes cross-expert low-frequency semantic agreement using a low-resolution view, thereby sidestepping optimization conflicts with established knowledge.
Extensive experiments on CIFAR-LT, ImageNet-LT, and iNaturalist 2018 show that VICAL consistently outperforms state-of-the-art methods, validating the effectiveness of our consistency-driven design.
Our code is available at \href{https://github.com/FlamieZhu/Vicinal-Consistency-Alignment}{VICAL}.
\end{abstract}

\section{Introduction}
\label{sec:intro}
Computer vision has witnessed remarkable progress over the last decade, largely driven by the construction of large-scale and high-quality datasets such as ImageNet~\cite{imagenet} and MS COCO~\cite{coco}. 
However, many of these benchmarks implicitly assume a balanced label distribution, which does not reflect real-world data distributions.
In practice, data typically follow a long-tailed distribution, where a few head classes dominate the data volume while the majority of tail classes suffer from extreme data scarcity.
Deep neural networks trained on such data exhibit a pronounced bias toward the head classes~\cite{bbn, decouple}, limiting applicability in domains such as medical diagnostics~\cite{medical} and autonomous driving~\cite{autonomous_driving}.

\begin{figure*}[t]
    \centering
    \includegraphics[width=1.0\linewidth]{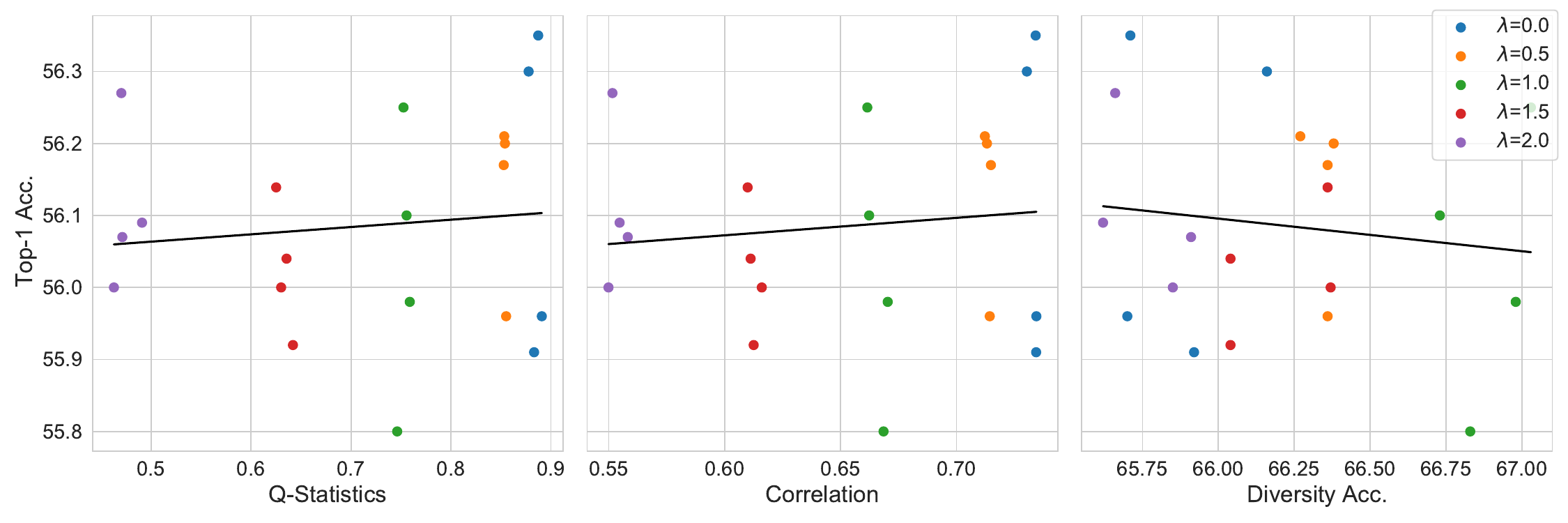}
    \caption{We use three different diversity metrics, \textit{i.e.}, Q-statistics, correlation, and diversity accuracy, to measure the relationship between model diversity and ensemble accuracy on CIFAR-100-LT. Each point denotes an independently trained model, and $\lambda$ controls the differences in logit adjustment strength across experts. 
    A larger $\lambda$ indicates more diverse experts by controlling the logit adjustment intensity for each expert. Varying $\lambda$ changes expert agreement but produces no clear improvement in ensemble accuracy. (The Pearson correlation coefficients are $0.11$, $0.11$, and $-0.13$ for Q-statistics, correlation, and diversity accuracy.)}
    \label{fig:div-acc}
\end{figure*}

To address the imbalance, numerous approaches have been proposed~\cite{bbn,decouple,ldam,bcl,paco,logit_adjust}.
Among them, logit adjustment~\cite{logit_adjust} is an effective approach to adjust the classification boundary and achieve Fisher consistency.
More recently, multi-expert models~\cite{ride,balpoe,mdcs,sade} have received significant attention due to their superior performance. 
This success is largely attributed to their ability to simultaneously mitigate both model variance and bias~\cite{ride}, thereby reducing prediction uncertainty~\cite{ecl}.
Some works~\cite{sade,balpoe,mdcs} also hypothesize that this success derives from model diversity, which maximizes expert specialization by forcing each expert to focus on different data distributions.
These approaches have achieved strong empirical performance, advancing long-tailed visual recognition.

However, the fundamental premise of these diversity-driven methods, \emph{the relationship between explicit expert diversity enforcement and ensemble prediction accuracy}, remains systematically unverified.
In this paper, we re-evaluate this core assumption using multiple ensemble diversity metrics and surprisingly reveal that expert diversity induced by logit adjustment or explicit loss regularizers does not universally yield better ensemble accuracy.
More formally, the expected test error for ensemble models can be approximately decomposed as~\cite{ensemble_diversity}
\begin{equation}\label{eq:ens_decom}
    Error = Noise +Bias+Variance-Diversity
\end{equation}
Eq.~\ref{eq:ens_decom} states diversity is an entangled hidden dimension in the bias-variance decomposition of ensemble learning rather than an independent variable~\cite{ensemble_diversity}.
Current expert diversity-enforcing methods~\cite{mdcs,ride} in long-tailed learning may overlook this crucial coupling.
As shown in Fig.~\ref{fig:div-acc}, we can obtain diverse outputs with lower Q-statistics and correlation coefficients by controlling the logit adjustment intensities for each expert, but without a notable improvement in ensemble accuracy.
We provide results on diversity induced by explicit regularizers\cite{ride} in the Supplementary Material, which show that explicitly forcing diversity degrades individual expert representation quality.
These observations prompt us to reconsider the underlying factors that truly determine the effectiveness of multi-expert models. 

Guided by this insight, we propose the Vicinal Consistency Alignment framework, VICAL, for \textit{variance reduction} rather than
\textit{explicit expert diversity constraints}. 
Our approach is grounded in a key insight: a robust model should exhibit minimal prediction variation against both localized high-frequency perturbations and global resolution shifts, especially for tail classes.
Specifically, VICAL comprises two complementary components operating across decoupled frequency domains:
Self-Consistency (SC) Learning and Deep Ensemble Distillation (DED).
First, SC enforces intra-expert prediction invariance by constructing a localized vicinity through the interpolation of strong augmented views.
By aligning predictions between the original views and these heavily corrupted interpolations, SC penalizes the exploitation of unstable high-frequency patterns, effectively smoothing the local loss landscape of tail classes.
Second, DED promotes inter-expert consistency by utilizing a low-resolution view to extract resolution-invariant semantics.
Because this downsampled input inherently lacks high-frequency details, DED strictly confines the cross-expert alignment to low-frequency semantic agreement.
This decoupling averts optimization conflicts, preserving the robust, individual features established by SC while filtering conflicting knowledge.
Together, these components significantly reduce prediction variance both within and across experts.

Our main contributions can be summarized as follows:
\begin{itemize}
    \item 
    Our work first empirically re-evaluates the hypothesis that expert diversity induced by logit adjustment or explicit regularizers directly correlates with ensemble accuracy in long-tailed recognition. Our study reveals that utilizing logit adjustment or explicit regularizers to maximize diversity can compromise individual expert accuracy, thereby undermining the overall ensemble.
    \item 
    We propose Vicinal Consistency Alignment (VICAL), a novel consistency learning framework that moves beyond explicit diversity constraints to implicit variance reduction. 
    \item 
    We implement VICAL through two core and complementary components: Self-Consistency Learning, which enables individual experts to discourage reliance on unstable high-frequency patterns, and Deep Ensemble Distillation, which facilitates cross-expert alignment through low-frequency semantic agreement.
    \item 
    VICAL showcases its superiority and generalizability through extensive experiments on popular long-tailed benchmarks, including CIFAR-LT, ImageNet-LT, and iNaturalist 2018, achieving significant gains over existing state-of-the-art methods.
\end{itemize}

\section{Related Work}
\noindent \textbf{Long-Tailed Visual Recognition} 
Conventional strategies for mitigating class imbalance primarily include re-sampling, re-weighting, and logit adjustment.
Re-sampling~\cite{ldam, decouple} techniques typically over-sample instances of tail classes or under-sample head classes, while re-weighting~\cite{cbloss,ldam} typically assigns higher loss weights to tail classes. Logit adjustment~\cite{logit_adjust,b_softmax} further refines this idea by explicitly modifying decision boundaries to accommodate class imbalance.
On the other hand, a large body of contrastive learning methods~\cite{paco,bcl, proco} has been adopted to enhance representation quality by learning a balanced feature space.

\noindent \textbf{Multi-Expert Models}
Recently, multi-expert models~\cite{ride,ncl,ncl++,sade,mdcs,balpoe} have become the dominant paradigm in long-tailed recognition. These methods typically train multiple experts independently or cooperatively, and aggregate predictions of all experts during testing.
A common hypothesis underpinning these approaches is that encouraging expert diversity can improve ensemble performance by allowing each expert to specialize in different categories. 
For example, RIDE~\cite{ride} enforces diversity by penalizing the similarity of expert predictions using the explicit diversity regularizer. SADE~\cite{sade}, MDCS~\cite{mdcs}, and BalPoE~\cite{balpoe} use varying logit adjustment intensities across experts, compelling each expert to specialize in distinct categories.
However, the validity of this foundational premise has rarely been scrutinized. Our work initiates from this critical gap.

\noindent \textbf{Consistency Learning} 
Consistency learning is a powerful regularization technique in semi-supervised learning~\cite{meanteacher,mdcs,fixmatch}, typically leveraging unlabeled data by enforcing output stability under input perturbations.
These perturbations are commonly generated through varying data augmentation strategies, such as the strong-weak pair employed in FixMatch~\cite{fixmatch}.
Beyond FixMatch, Mean Teacher~\cite{meanteacher} also uses generic consistency to propagate pseudo-labels to unlabeled data.
Consistency learning has also been shown to improve model robustness and representation learning~\cite{augmix}. 
Mixup~\cite{mixup} creates vicinal samples by mixing instances from different classes, yielding more robust decision boundaries.
GLMC~\cite{glmc} utilizes Mixup and CutMix~\cite{cutmix} to achieve global and local mixture consistency for long-tailed learning.
In contrast, VICAL adopts frequency-decoupled consistency, which acts as a spectral bottleneck to suppress high-frequency overfitting and enforce a robust, low-frequency consensus. This design reduces variance within and across experts in long-tailed recognition.

\section{Do More Diverse Experts Yield Better Performance?}
A line of recent methods~\cite{sade, mdcs, ace,balpoe,ssl_expert,prl} has been proposed to improve the output diversity of multi-expert models by enabling each expert to focus on different categories.
However, one premise for these practices to be effective is that diverse experts are beneficial for long-tailed learning, and this simple assumption has not been verified.
In this section, we first provide preliminaries for long-tailed recognition and then evaluate whether improving the expert diversity through logit adjustment and an explicit regularizer in existing long-tailed learning approaches enhances recognition performance.

\subsection{Preliminaries}
The goal of long-tailed visual recognition is to train on skewed data while achieving promising performance on the balanced test set.
Formally, let $\mathbb{D}=\{ x_{i},y_{i} \}_{i=1}^{N}$ be the training set where $x_{i}$ denotes the $i$-th image sample and $y_{i}$ is the corresponding label. For an ensemble model with $M$ experts, given the input $x_{i}$, the prediction probability of $j$-th class in $m$-th expert parameterized with $\theta_{m}$ is 
\begin{equation}\label{eq:prob_ori}
    p_{j}(x_{i};\theta_{m})=\frac{exp(z^{m}_{ij})}{\sum_{k=1}^{C}exp(z^{m}_{ik})},
\end{equation}
where $z^{m}_{ij}$ is the logit for class $j$ of sample $i$, and $C$ is the number of classes.
The posterior in Eq.~\ref{eq:prob_ori} does not account for the train-test class prior shift. After calibration by logit adjustment~\cite{logit_adjust,b_softmax}, Eq.~\ref{eq:prob_ori} can be reformulated as 
\begin{equation}\label{eq:prob_la}
    p_{j}(x_{i};\theta_{m})=\frac{exp(z^{m}_{ij}+\tau_m\log(P_{j}))}{\sum_{k=1}^{C}exp(z^{m}_{ik}+\tau_m\log(P_{k}))},
\end{equation}
\begin{equation}
    \mathcal{L}_{ce}^{m} = -\log(p_{j}(x_{i};\theta_{m})).
\end{equation}
Here, $P_{j}$ is the class prior of class $j$, the hyperparameter $\lambda$ controls calibration strength, and $\mathcal{L}_{ce}^{m}$ denotes the basic classification loss of individual experts.
A large value of $\tau_m$ ($>$1) will render the prediction biased towards the tail class and a small value ($<$1) towards the head class. 
Previous works~\cite{mdcs, balpoe, ssl_expert, sade} assign small and large $\lambda$s for head and tail experts, focusing on head and tail classes, respectively. 
Following the common setup~\cite{balpoe,mdcs,sade}, we set $M=3$ in subsequent sections unless otherwise mentioned.

\subsection{Statistical Analysis}
In our work, we find that expert diversity induced by logit adjustment (See Fig.~\ref{fig:div-acc}) or an explicit loss regularizer (See Supplementary Material) has no positive correlation with ensemble performance.
Specifically, we train multi-expert models on CIFAR-100-LT (IF=100) under varying logit adjustment intensities: $(\tau_1,\tau_2,\tau_3)=(1-\lambda$, $1$, $1+\lambda)$ ($\lambda \in \{0, 0.5, 1, 1.5, 2\}$) 
for the long-tailed, uniform, and reversed experts, respectively.
We measure diversity using Q-statistics~\cite{q-statistics} and the correlation coefficient $\rho$, with the diversity factor $\sigma$~\cite{mdcs} as an auxiliary accuracy metric.
As shown in Fig. \ref{fig:div-acc}, larger $\lambda$ values yield lower Q-statistics and $\rho$, indicating that more diverse predictions are obtained. 
However, this induced diversity fails to yield ensemble accuracy gains.
Furthermore, when applying explicit regularizers (See Supplementary Material), we observe a strong positive correlation between Q-statistics and ensemble accuracy.
This indicates that artificially pushing experts to be diverse harms overall performance, motivating our shift in focus from pursuing diversity to reducing model variance.

\section{Vicinal Consistency Alignment}
In this paper, we emphasize that reducing the variance of the multi-expert model is more effective than enforcing diversity in Eq.~\ref{eq:ens_decom}.
To this end, we present a complementary and tiered consistency-learning framework, VICAL, to achieve vicinal consistency alignment, with two key components: Self-Consistency (SC) Learning and Deep Ensemble Distillation (DED).
Specifically, we construct a vicinity for each instance using an interpolation of two strongly augmented full-resolution views, together with a separate low-resolution view.
The vicinity is strictly confined to transformations of the same input. 
First, SC acts as the local smoother and stabilizes the individual expert's prediction within the vicinity of each sample. 
Second, DED aligns the individual output of the low-resolution view with the global semantic consensus based on SC.
Notably, by separating these objectives across different input resolutions, VICAL successfully reduces model variance both within and across experts while avoiding knowledge conflicts.
In the following section, we will describe our proposed VICAL in greater detail.

\subsection{Self-Consistency Learning}
\textbf{Vanilla Self-Distillation} Self-distillation (SD) is an elegant and efficient approach to reducing model variance and improving prediction consistency~\cite{ban,byot,meanteacher}. Here, we consider an exponential moving average (EMA) model as the teacher to produce soft labels, which are more informative than hard labels.
For simplicity, let $p^{\mathcal{S}}$ and $p^{\mathcal{T}}$ denote the predicted probability of the student and teacher model, respectively.
The vanilla SD~\cite{meanteacher} of the $i$-th sample in the $m$-th expert can be expressed as:
\begin{equation}
\begin{aligned}
    \mathcal{L}_{sd} = KL(p^{\mathcal{T}}(x_{i};\hat{\theta}_{m}) || p^{\mathcal{S}}(x_{i};\theta_{m})), 
\end{aligned}
\end{equation}
where $KL(\cdot || \cdot)$ represents the Kullback–Leibler divergence and $\hat{\theta}_{m}$ denotes the EMA model of $m$-th expert, updated from the online model $\theta_{m}$. We omit the parameters $\theta_{m}$ and $\hat{\theta}_{m}$ in the following sections for simplicity.

\begin{figure}[t] 
    \centering
    \subfloat[]{\includegraphics[scale=0.275]{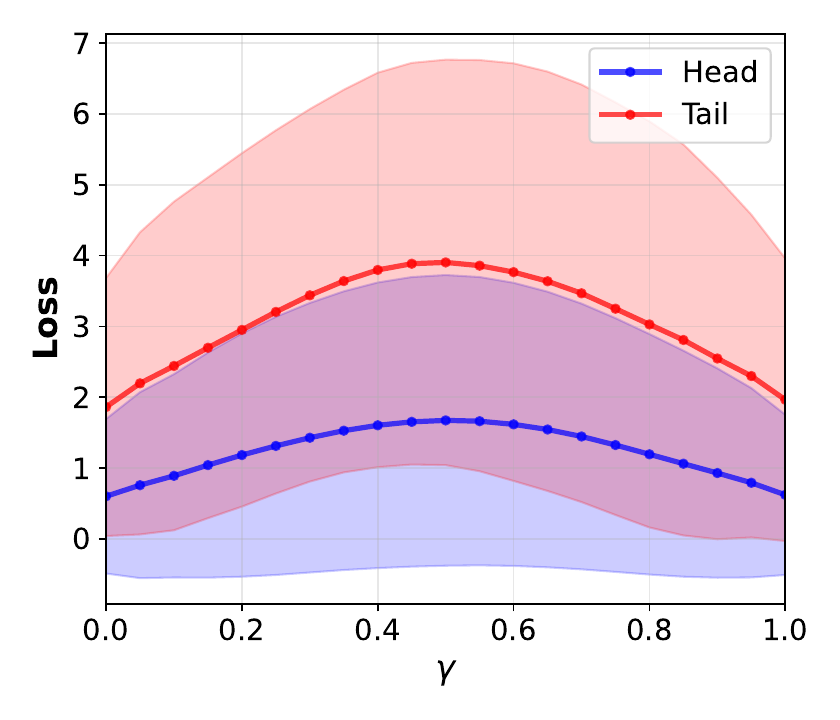} \label{fig: training_loss_base} }
     \subfloat[]{\includegraphics[scale=0.275]{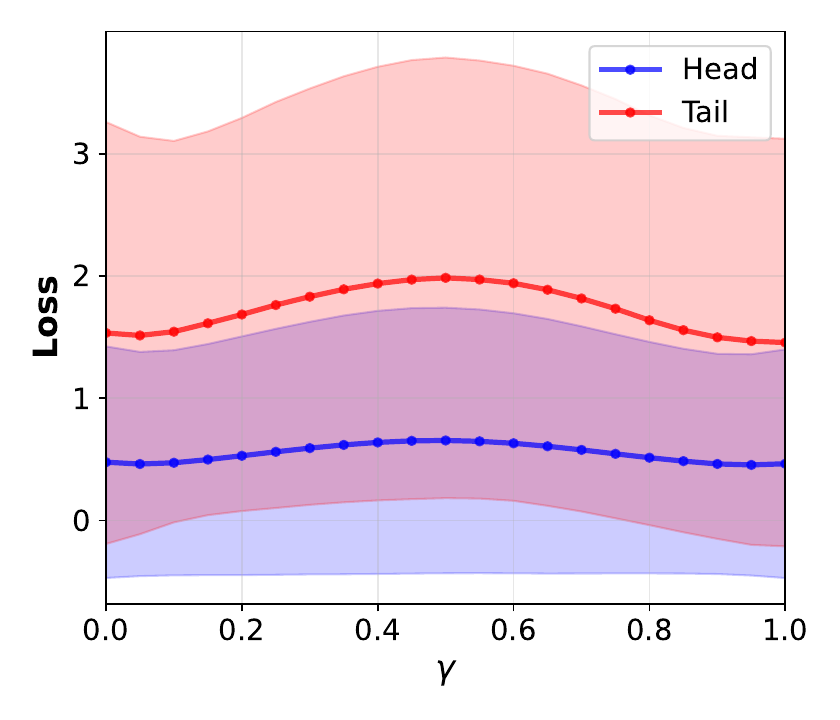} \label{fig: training_loss_best}} 
     \subfloat[]{\includegraphics[scale=0.365]{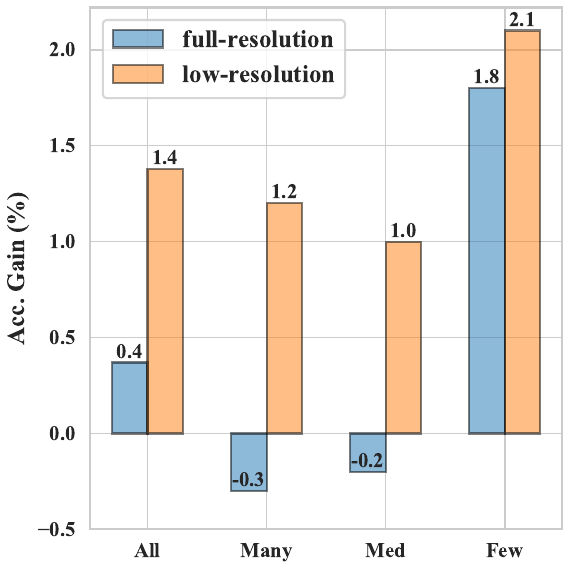} \label{fig: ablation_res}} 
    \caption{
    Loss landscapes as a function of the interpolation factor $\gamma$ $\in$ [0,1] between two augmented views $\mathrm{v_{1}}$ and $\mathrm{v_{2}}$ on CIFAR-100-LT (IF=100) with a single expert: (a) \textbf{vanilla self-distillation}, (b) \textbf{SC}. SC induces a smoother loss curve as $\gamma$ varies, especially for the tail classes.
    (c) Accuracy gains over the SC baseline when DED uses a full-resolution or low-resolution input.
    }
    \label{fig: training_loss}
\end{figure}

\textbf{Inconsistent Loss Landscape} 
We find vanilla SD inadequate for enforcing robustness within the vicinity of each training sample. 
Specifically, we construct an interpolated view $\widetilde{\mathrm{v}}=\gamma \mathrm{v}_{1}+ (1-\gamma)\mathrm{v}_{2}$ within the vicinity, where $\mathrm{v}_{1}$ and $\mathrm{v}_{2}$ are two strong augmentations of the same image $x$, and $\gamma \sim Beta(\alpha, \alpha)$, with $\gamma \in[0,1]$.
Since $\mathrm{v}_{1}$ and $\mathrm{v}_{2}$ are semantically correlated, the loss of the convex combination $\widetilde{\mathrm{v}}$ for the tail class is expected to be as smooth as that of the head class when varying with $\gamma$.
However, as shown in Fig.~\ref{fig: training_loss_base}, the vanilla SD exhibits high sensitivity to $\gamma$ for the tail class, suggesting a sharp loss landscape and substantial prediction uncertainty.

\begin{figure*}[t]
    \centering
    \includegraphics[width=1.0\linewidth]{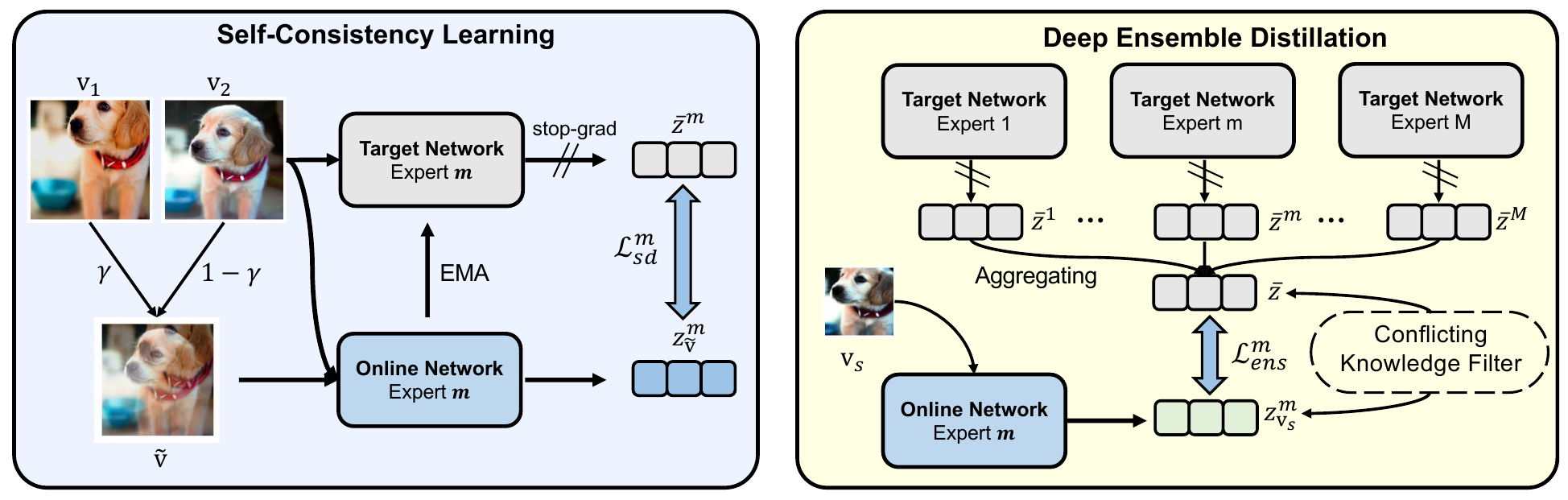}
    \caption{The framework of our proposed VICAL, consisting of Self-Consistency (SC) Learning and Deep Ensemble Distillation (DED). SC utilizes the interpolated view $\widetilde{\mathrm{v}}$ to discourage reliance on unstable high-frequency patterns, significantly improving robustness of tail classes.
    DED strictly confines the cross-expert alignment to low-frequency semantic agreement via a low-resolution view $\mathrm{v}_{s}$, avoiding optimization conflicts with the SC objective.
    DED further filters conflicting transfers while retaining complementary expert knowledge through the Conflicting Knowledge Filter.
    }
    \label{fig:framework}

\end{figure*}

\textbf{Local and High-frequency Perturbations} 
To mitigate sharp loss variations within the vicinity, we propose a self-consistency (SC) objective as the local smoother that penalizes prediction divergence across interpolated views.
Specifically, we employ two strong augmented views $\mathrm{v}_{1}$ and $\mathrm{v}_{2}$,  and introduce unstable high-frequency perturbation via the interpolated view $\widetilde{\mathrm{v}}$, which creates unnatural edges, ghosts, and textural artifacts. 
By feeding this heavily corrupted view $\widetilde{\mathrm{v}}$ into the online student network, we compel the model to align its prediction with the EMA teacher of the original views.
Due to data scarcity, tail classes are highly sensitive to vicinal perturbations, thereby receiving stronger regularization during consistency alignment.
Consequently, SC penalizes the exploitation of unstable high-frequency patterns and achieves prediction consistency within the vicinity for each expert. By suppressing these brittle features, SC alleviates overfitting in the tail classes and yields flatter local minima (See Fig.~\ref{fig: training_loss_best}).

As illustrated in Fig.~\ref{fig:framework}, the interpolated view $\widetilde{\mathrm{v}}$ is fed into the online network to produce the prediction $p^{\mathcal{S}}(z^{m}_{\widetilde{\mathrm{v}}})$ of the $m$-th expert as the student distribution.
The views $\mathrm{v}_{1}$ and $\mathrm{v}_{2}$ are fed into the target network to calculate the mean logits $\overline{z}^{m}$ to serve as the stable teacher.
Formally, the modified objective function of the $m$-th expert can be expressed as

\begin{equation}\label{eq:sd}
\begin{aligned}
    \mathcal{L}^{m}_{sd} =  KL({p}^{\mathcal{T}}(\overline{z}^{m}) || p^{\mathcal{S}}(z^{m}_{\widetilde{\mathrm{v}}})) , 
\end{aligned}
\end{equation}
where $\overline{z}^{m}= \frac{1}{2}(z^{m}_{\mathrm{v}_{1}}+z^{m}_{\mathrm{v}_{2}} )$ is the mean of EMA logits for $\mathrm{v_{1}}$ and $\mathrm{v_{2}}$ and serves as the stabilized target.

\subsection{Deep Ensemble Distillation}
\textbf{Global and Low-frequency Alignment} 
Ensemble distillation offers a promising approach for variance reduction~\cite{ride}.
While SC improves robustness to high-frequency perturbations introduced by interpolation, unconditionally forcing the student to align with the global ensemble consensus poses a severe risk of optimization conflict with established knowledge of SC.
Concretely, if an extra full-resolution view were employed for distillation,  ensemble averaging inherently blurs individual features established by SC, leading to degraded performance compared to using SC alone (See Fig.~\ref{fig: ablation_res}).
To sidestep this conflict, our Deep Ensemble Distillation (DED) module restricts the distillation objective strictly to a low-resolution view $\mathrm{v}_{s}$.
Because the downsampled $\mathrm{v}_{s}$ lacks high-frequency information, this resolution asymmetry confines the ensemble alignment strictly to low-frequency semantics.
Consequently, the DED module offers complementary signals based on SC, and our design decouples the SC and DED objectives across different frequency domains.

As shown in Fig.~\ref{fig:framework}, the low-resolution view $\mathrm{v}_{s}$ is fed into the online network and produces logits $z^{m}_{\mathrm{v}_{s}}$ of the $m$-th expert.
The outputs of all target models are averaged to obtain $\overline{z}$, which provides the global semantic consensus. The logits $z^{m}_{\mathrm{v}_{s}}$ are forced to align with the mean $\overline{z}$ of full-resolution views, which enhances low-frequency consistency and improves overall robustness.

\textbf{Conflicting Knowledge Filter}
To further ensure that each expert gains complementary knowledge, we introduce the Conflicting Knowledge Filter (CKF) that sidesteps incorrect knowledge transfer, preventing individual knowledge from degenerating into uncertain global consensus.
The final ensemble distillation formulation can be expressed as 
\begin{equation}\label{eq:ded}
    \mathcal{L}^{m}_{ens} =  \frac{1}{|\mathbb{D}^{C}_{c}|}\sum_ {\mathrm{v}_{s}\in \mathbb{D}^{C}_{c}}KL({p}^{\mathcal{T}}(\overline{z}) || p^{\mathcal{S}}(z^{m}_{\mathrm{v}_{s}})) ,
\end{equation}
\begin{equation}\label{eq:ckf}
\small
    \mathbb{D}_{c} = \{ x | \mathrm{argmax} ({p}^{\mathcal{T}}(\overline{z})) \neq y\ and \ 
      \mathrm{argmax}({p}^{\mathcal{S}}(z^{m}_{\mathrm{v}_{s}})) = y \}
\end{equation}
where $\overline{z}$ denotes the aggregated logits across experts, and $\mathbb{D}^{C}_{c}$ denotes the complement of $\mathbb{D}_{c}$, \textit{i.e.}, the non-conflict knowledge set, excluding samples that the student correctly predicts but the teacher fails. 
The CKF maintains individual experts' specialized knowledge while enforcing semantic-level agreement.

\begin{table*}[t]
    \setlength{\tabcolsep}{3.5pt}
    \centering
    \begin{tabular}{l|cccc|cccc|c}
     \toprule
     \multirow{2}{*}{Method} & \multicolumn{4}{c|}{CIFAR-100-LT} & \multicolumn{4}{c}{CIFAR-10-LT} &
     \multicolumn{1}{c}{ImageNet-LT}
     \\
     
    & 200 &100 & 50 & 10 & 200 & 100 & 50 & 10 & 256\\
     \midrule
     
     LDAM-DRW \cite{ldam} & 38.5& 42.0 & 46.6 & 58.7 & 74.7&77.0 & 81.0 & 88.2 & 45.8 \\
     
     BCL$^{\ddag}$\cite{bcl} & - & 51.9 & 56.6 & 64.9 &-& 84.3 & 87.2 & 91.1 & 57.1\\
     PaCo$^{\ddag}$\cite{paco} &  47.8& 52.0& 56.0& 64.2 & 82.3&85.4&88.0  & 91.5& 57.2   \\
     ProCo$^{\ddag}$\cite{proco} &  -&52.8& 57.1& 65.5 & - &85.9& 88.2 & 91.9 & 58.0  \\
     
     \midrule
     RIDE \cite{ride} & 44.6& 48.0 & 51.7 & 61.8 & 77.8 & 81.2 & 83.7&86.3 & 56.8\\
     SADE\cite{sade} &44.7& 49.8 & 53.9 & 63.6 & 78.0 & 82.9 & 85.8&90.0 & 58.8 \\

     NCL$^{\ddag}$\cite{ncl}& 49.5 & 54.2 & 58.2 & 63.8 & 82.2 & 85.5& 87.3& 91.1 & 60.5  \\
     
       BalPoE$^{\ddag}$\cite{balpoe}& -&55.9&60.1 & 68.1 & - &86.8 &88.5& \underline{91.9} & 61.6\\
     MDCS$^{\ddag}$\cite{mdcs}& -&56.1& 60.1&-&-& 87.2&88.3&- & 60.2\\
     ECL$^{\ddag}$\cite{ecl}& \underline{51.4} & 56.3 & 59.9 & 67.3 & \underline{83.6} & 86.5& 88.9& 91.8 & \underline{61.7} \\
     PRL\cite{prl} & - & 52.8 & 57.3 & 65.6&- & - & - & - & 60.7\\
     
     NCL++$^{\ddag}$\cite{ncl++} &  - & 56.3 & 59.8 & - & - & 87.2 & 88.8 & - & 60.9\\
     ICL$^{\ddag}$\cite{icl} &  - & \underline{57.6} & \underline{61.3} & \underline{69.3} & - & \underline{87.9} & \underline{89.7} & \underline{91.9} & 60.2\\
     \midrule
     
     \rowcolor{yellow!30}
     \textbf{Ours}$^{\ddag}$ & \underline{\textbf{55.3}} & \underline{\textbf{59.7}} & \underline{\textbf{63.9}} &
     \underline{\textbf{71.6}} &
     \underline{\textbf{86.7}} &
     \underline{\textbf{90.0}} &
     \underline{\textbf{91.9}} &
     \underline{\textbf{94.5}} &
     \underline{\textbf{62.9}} 
     \\ 
     \bottomrule
\end{tabular}
\caption{Top-1 accuracy on CIFAR-LT using ResNet-32 as backbone and on ImageNet-LT using ResNeXt-50 as backbone. 
We report the results of 400 and 200 epochs for CIFAR-LT and ImageNet-LT, respectively. $\ddag$ denotes models trained with strong augmentation\cite{paco}. The best result is shown in bold, and the second-best result is underlined.}
    \label{tab:cifar-lt}
\end{table*}

\subsection{Training and Inference}
\textbf{Training} 
The overall training objective comprises three components: cross-entropy loss $\mathcal{L}^{m}_{ce}$ of original views $\mathrm{v}_{1}$ and $\mathrm{v}_{2}$, self-consistency distillation $\mathcal{L}^{m}_{sd}$ of $\widetilde{\mathrm{v}}$, and ensemble distillation loss of $\mathcal{L}^{m}_{ens}$ of $\mathrm{v}_{s}$. Finally, the loss is as follows:  

\begin{equation}
    \mathcal{L}= \sum_{m=1}^{M}
    \left(
    \mathcal{L}^{m}_{ce}+\eta \mathcal{L}^{m}_{ens}+\beta\mathcal{L}^{m}_{sd}\right),
\end{equation}
where $\eta$ and $\beta$ are two hyperparameters to control the distillation strength, respectively.
Note that each expert is trained with a uniform-aware objective to produce class-balanced predictions.

\noindent \textbf{Inference}  The online network's exposure to the interpolated samples degrades the precision of its Batch Normalization statistics. Consequently, we only keep the target network during the inference phase to produce high-quality predictions, and the online network is discarded.

\section{Experiments}
\label{sec: experiments}
To validate the efficacy of our proposed VICAL, we conduct comprehensive experiments on long-tailed CIFAR-10~\cite{ldam}, long-tailed CIFAR-100~\cite{ldam}, ImageNet-LT~\cite{oltr}, and iNaturalist 2018~\cite{inaturalist}. Details of these datasets and extended experiments are available in the Supplementary Material.

\begin{table}[t]
    \centering
    \begin{minipage}{0.485\textwidth}
    \setlength{\tabcolsep}{3.5pt}
    \centering
        \begin{tabular}{lccc|c}
        \toprule
        Method & Many & Med & Few & All\\
    
        \midrule
        100 epochs \\
        % BCL$^{\ddag}$\cite{bcl} & - & - & - & 71.8  \\
        RIDE\cite{ride} & 70.9 & 72.4 & 73.1 & 72.6 \\
        ACE\cite{ace} & - & - & - & 72.9 \\
        BalPoE\cite{balpoe} & 73.2 & 75.5 & 74.7 & 75.0 \\ 
        BalPoE$^{\ddag}$\cite{balpoe} & - & - & - & 73.5 \\
        MDCS$^{\ddag}$\cite{mdcs} & 71.8 & 73.1 & 72.4 & 72.5 \\
        \midrule
        \rowcolor{yellow!30}
        \textbf{Ours}$^{\ddag}$ &  \textbf{74.3} &  \textbf{77.3} & \textbf{76.2} & \textbf{76.6}   \\
        \midrule
        200 epochs \\
        TS-MOF\cite{tsmof} & 72.8 & 73.6 & 70.5 & 72.3  \\
        ResLT\cite{reslt} & 73.0 & 72.6 & 73.1 & 72.9 \\
        % SADE\cite{sade} & - & - & - & 72.9\\
        ML\cite{mutual} & - & - & - & 74.9\\
        % PRL\cite{prl} & - & - & - & 75.1 \\
        NCL++$^{\ddag}$\cite{ncl++} & 72.2 & 75.3 & 75.7 & 75.2 \\
        SHIKE$^{\ddag}$\cite{shike} & - & - & - & 75.4  \\
        ICL$^{\ddag}$\cite{icl} & 74.9 & 75.9 & 76.1 & 75.9\\
        \midrule
        \rowcolor{yellow!30}
        \textbf{Ours}$^{\ddag}$ &  \textbf{75.5} &  \textbf{78.1} & \textbf{77.3} & \textbf{77.5}   \\ 
        \bottomrule
        \end{tabular}
        \caption{Top-1 accuracy on iNaturalist 2018 using ResNet-50 as backbone. We report the results of 100 epochs and 200 epochs, respectively. $\ddag$ denotes models trained with RandAugment\cite{randaug}.}
        \label{tab:img_inat}
    \end{minipage}
    \hfill
    \begin{minipage}{0.485\textwidth}
    \setlength{\tabcolsep}{4pt}
    \centering
    \begin{tabular}{ccccc}
    \toprule
    \multirow{2}{*}{ME} & \multicolumn{1}{c}{SC} & \multicolumn{1}{c}{DED}  & \multicolumn{1}{c}{CKF}  & \multirow{2}{*}{Top-1 Acc.} \\
    & Eq.~\ref{eq:sd} & Eq.~\ref{eq:ded} & Eq.~\ref{eq:ckf}\\
    \midrule
    \checkmark & & &  & 55.6\\
    \checkmark & \checkmark & &  & 57.8\\
     \checkmark & \checkmark & \checkmark & &  58.6 \\
      \checkmark & \checkmark & \checkmark  & \checkmark & 59.1  \\
    \bottomrule
    \end{tabular}
    \caption{Ablation study on CIFAR-100-LT training for 250 epochs. ME: multi-expert model. SC: Self-Consistency Learning. DED: Deep Ensemble Distillation. CKF: Conflicting Knowledge Filter.}
    \label{tab:ablation_cifar}
    \begin{tabular}{lc}
    \toprule
    Methods & Top-1 Acc. \\
    \midrule
    full-resolution & 58.1 \\
     + high-pass & 58.1 \\
     + low-pass & 58.6 \\
    \midrule
    low-resolution & 59.1 \\ 
    \bottomrule
    \end{tabular}
    \caption{Ablation study of applying spatial frequency filters (high-pass vs. low-pass) to the full-resolution input images for DED on CIFAR-100-LT.}
    \label{tab:ablation_resolution_freq}
    \end{minipage}
\end{table}

\subsection{Experimental Settings}
\noindent \textbf{Evaluation Protocol} 
We follow the standard evaluation protocol~\cite{ldam,bbn,paco,bcl} and validate the top-1 accuracy for all categories on the corresponding balanced test set. 
Following common practice~\cite{decouple}, we split the classes into three groups, \textit{i.e.}, many-shot ($>$100 images), medium-shot (20$\sim$100 images), and few-shot ($<$20 images), and report the respective accuracy of each group. 

\noindent \textbf{Implementation Details} 
We implement our approach with PyTorch~\cite{pytorch} and employ the SGD optimizer with a momentum of 0.9 by default.
The models are trained for 400, 200, and 100/200 epochs, and the batch size is set to 64, 256, and 512 for CIFAR-LT, ImageNet-LT, and iNaturalist 2018, respectively.
For network architecture and data augmentation strategy, we mainly follow \cite{balpoe, mdcs}.
The initial learning rate is 0.2 for iNaturalist 2018 and 0.1 for the other datasets. The learning rate follows a cosine annealing schedule for ImageNet-LT and iNaturalist 2018, and is multiplied by 0.1 at epochs 320 and 360 for CIFAR-LT.
We set $\alpha$ to 1.0, $\eta$ to 0.8, and $\beta$ to 1.0 for all experiments. 
The momentum coefficient of the target network is set to 0.99 by default.
The resolution of the low-resolution view is set to 16$\times$16 for CIFAR-LT and 96$\times$96 for the large-scale datasets.
More details can be found in the Supplementary Material.

\begin{table*}[t]
    \centering
    \resizebox{\textwidth}{!}{
    \begin{tabular}{l|c|c|c|c|c}
    \toprule
    \textbf{Methods} & \textbf{Multi-Expert} & \textbf{Intra-Expert Stabilization} & \textbf{Inter-Expert Alignment} & \textbf{Objective Decoupling} & \textbf{Distillation Gate} \\
    \midrule
    PCL\cite{pred_consistency} & $\times$ & Predictive Consistency & $\times$ & $\times$ & $\times$ \\
    GLMC\cite{glmc} & $\times$ & Inter-Class Mixup & $\times$ & $\times$ & $\times$ \\
    LTRL\cite{ltrl} & $\times$ & Reflective Learning & $\times$ & $\times$ & $\times$ \\
    RIDE\cite{ride} & \checkmark & $\times$ & Explicit Divergence & $\times$ & $\times$ \\
    BalPoE\cite{balpoe} & \checkmark & Inter-class Mixup & Explicit Divergence & $\times$ & $\times$ \\
    MDCS\cite{mdcs} & \checkmark & Vanilla Self-Distillation & Explicit Divergence & $\times$ & 
    Confident Sampling\\
    NCL++\cite{ncl++} & \checkmark & Vanilla Self-Distillation & Full-Resolution Mimicry & $\times$  & $\times$ (Unconditional) \\
    ICL\cite{icl} & \checkmark & $\times$ & Full-Resolution Mimicry & $\times$ & $\times$ (Unconditional) \\
    \midrule
    \rowcolor{yellow!30} \textbf{VICAL (Ours)} & \textbf{\checkmark} & \textbf{Vicinal Interpolation} & \textbf{Low-Resolution Semantic} & \textbf{Resolution Asymmetry} & \textbf{CKF} \\
    \bottomrule
    \end{tabular}
    }
    \caption{Mechanism comparisons of VICAL against existing consistency-driven and multi-expert methods. Unlike prior approaches that rely on explicit divergence or coupled full-resolution distillation, VICAL decouples intra-expert local smoothing from inter-expert global semantic agreement, which avoids optimization conflicts.}
    \label{tab:comparison_consis_ensemble}
\end{table*}

\subsection{Main Results}
\noindent \textbf{Comparison with state-of-the-art}
We compare VICAL with previous state-of-the-art methods such as NCL++\cite{ncl++}, MDCS\cite{mdcs} and BalPoE~\cite{balpoe}. 
BalPoE serves as a representative baseline that integrates Mixup during training.
Tab.~\ref{tab:cifar-lt} lists the results on CIFAR-LT and ImageNet-LT, and Tab.~\ref{tab:img_inat} shows the comparison on iNaturalist 2018. 
Our VICAL framework consistently achieves state-of-the-art performance across all datasets. 
Tab.~\ref{tab:comparison_consis_ensemble} lists the core mechanism differences of VICAL against existing consistency-driven and multi-expert approaches.
Compared to NCL++~\cite{ncl++} and MDCS~\cite{mdcs}, which employ full-resolution views for distillation, VICAL achieves substantial improvement, validating the efficacy of our vicinal consistency alignment.

\begin{figure}[t] 
    \centering
    \subfloat[]{\includegraphics[scale=0.39]{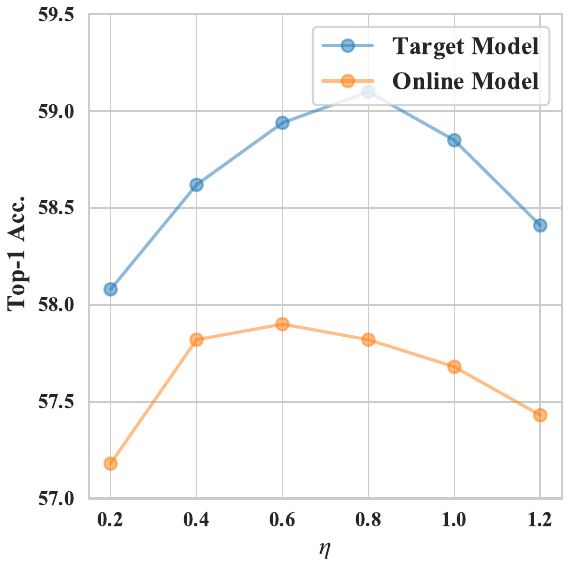} \label{fig: ablation_eta} }
     \subfloat[]{\includegraphics[scale=0.39]{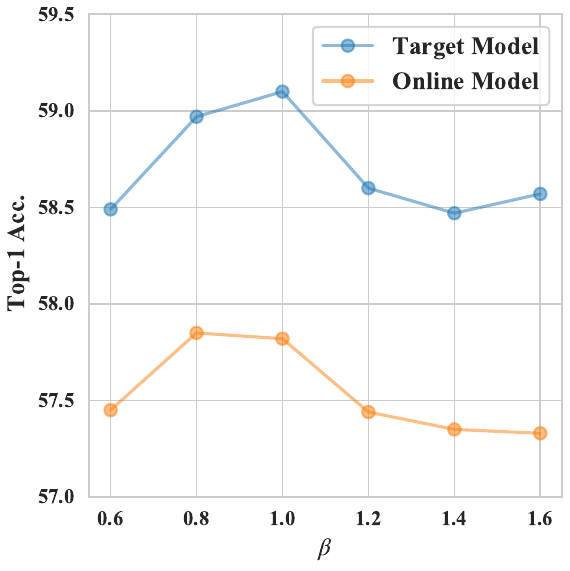} \label{fig: ablation_beta}}
     \subfloat[]{\includegraphics[scale=0.39]{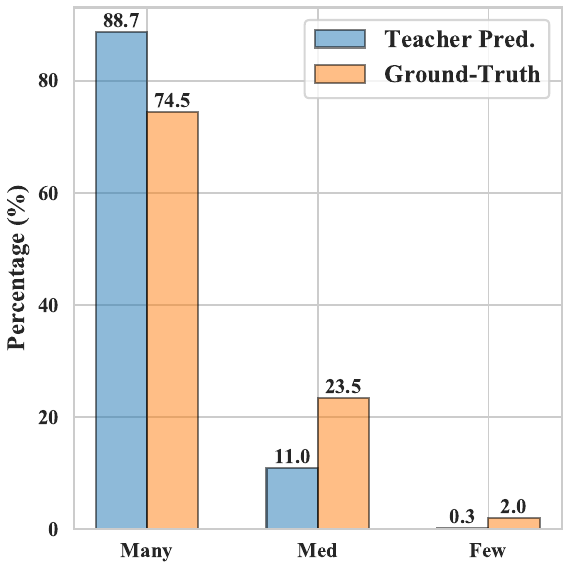}
     \label{fig: conflict_cls_analysis}}
    \caption{
    Ablation study of the hyperparameters (a) $\eta$ and (b) $\beta$ on CIFAR-100-LT (IF=100). We report the best results achieved by the online and target models. 
    (c) By the terminal training phase, the histogram of the conflicting knowledge set $\mathbb{D}_{c}$. 
    }
    \label{fig: ablation}
\end{figure}

\noindent \textbf{Component Analysis} 
Tab.~\ref{tab:ablation_cifar} presents the ablation study of all components of our VICAL on CIFAR-100-LT. The baseline achieves 55.6$\%$ top-1 accuracy. Each module provides a substantial improvement, indicating that the components are complementary: SC aligns predictions within each expert, while DED achieves low-frequency semantic agreement across experts.

\noindent \textbf{Effect of Hyperparameters} 
Fig.~\ref{fig: ablation_eta} and  Fig.~\ref{fig: ablation_beta} present the ablation studies of hyperparameters $\eta$ and $\beta$. 
The target model achieves the best accuracy at $\eta=0.8$ while the online network peaks at $\eta=0.6$. Similarly, the optimal value of $\beta$ for the target model is $1$.
The superior performance of the target model is attributed to its clean inputs. In contrast, exposure to corrupted samples introduces noise into the online network and degrades its Batch Normalization statistics.
Consequently, we use the target network during inference.

\begin{figure}[t] 
    \centering
    \subfloat[]{\includegraphics[scale=0.198]{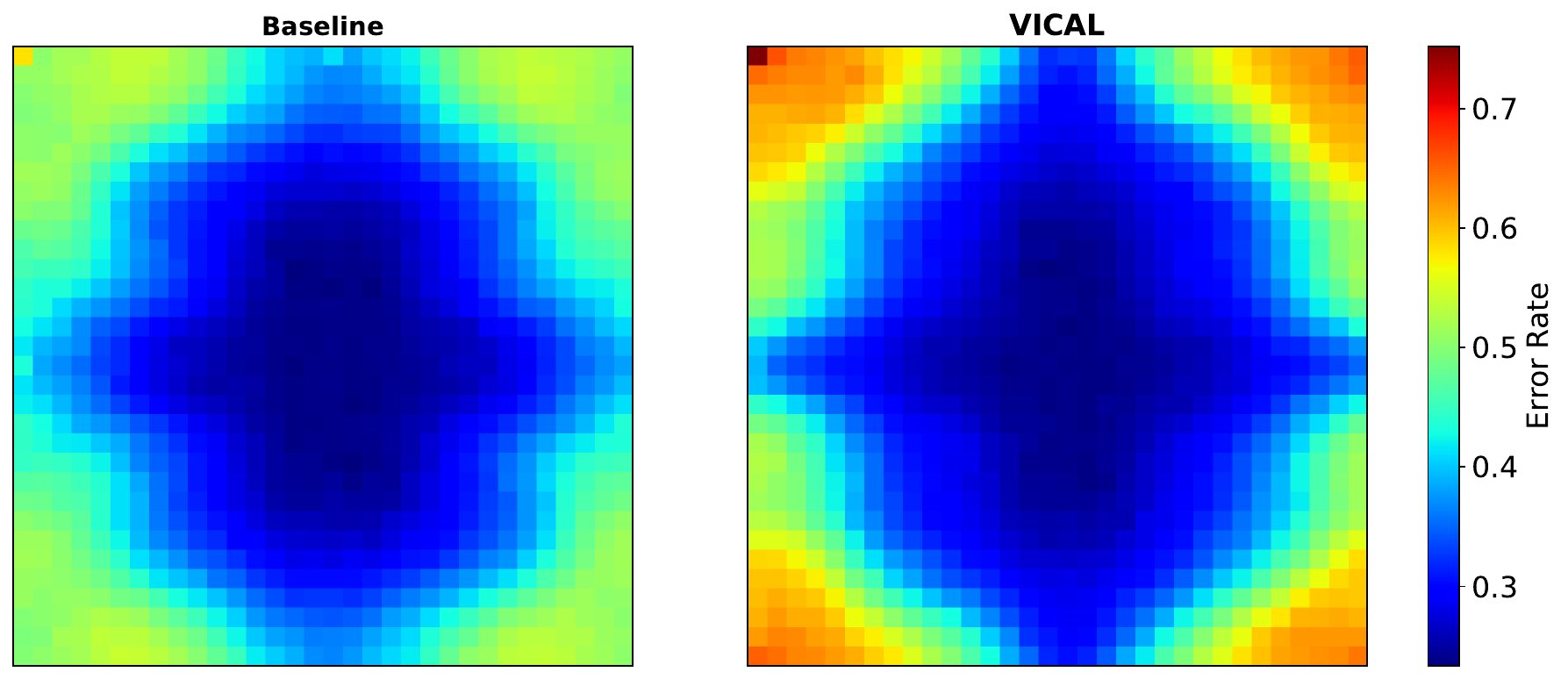} \label{fig: fourier_head} }
     \subfloat[]{\includegraphics[scale=0.198]{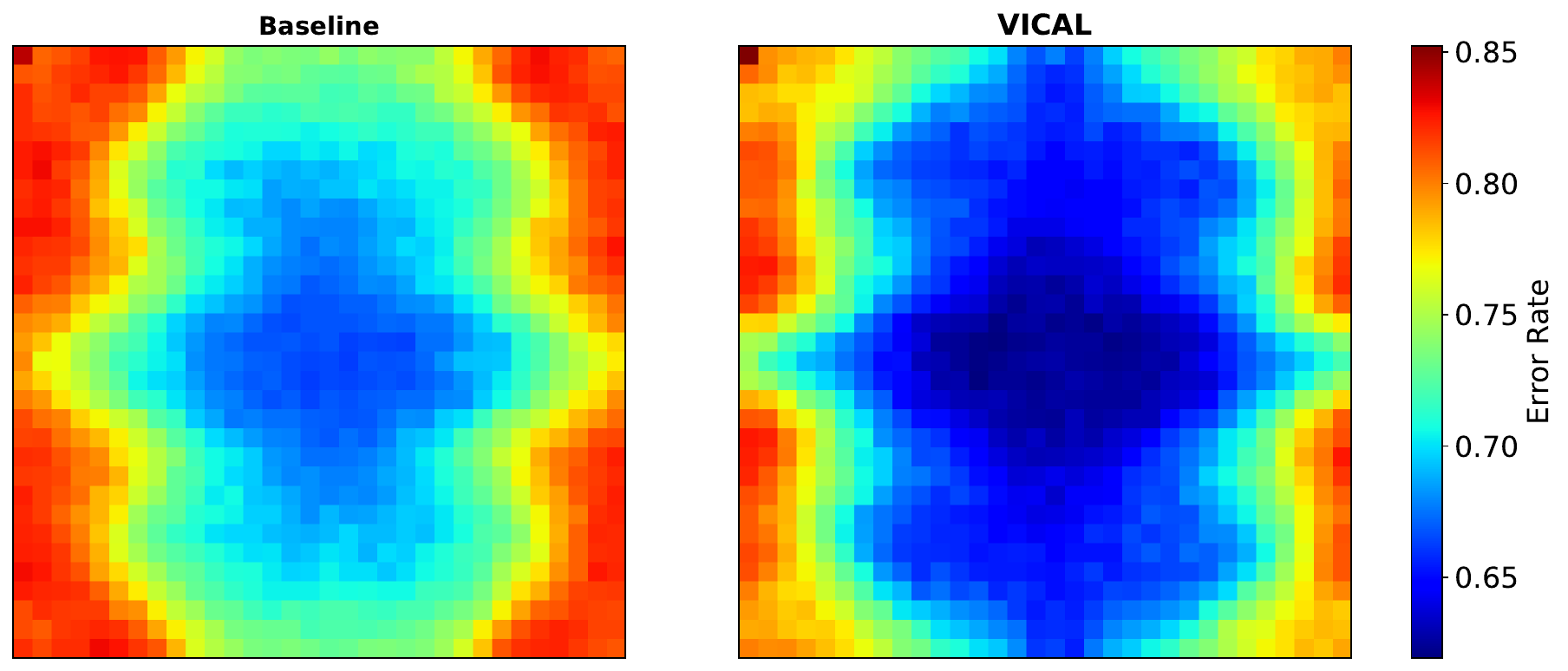} \label{fig: fourier_tail}}
    \caption{
    Model sensitivity to additive noise aligned with different Fourier basis vectors on the CIFAR-100-LT validation set: (a) \textbf{Head} and (b) \textbf{Tail}. The center represents \textbf{low-frequency}, while the edges represent \textbf{high-frequency}. For head classes, VICAL is slightly more sensitive to extreme high-frequency noise. In contrast, VICAL substantially improves tail class robustness in the low-frequency region.
    }
    \label{fig: fourier}
\end{figure}

\begin{table}[t]
    \centering
    \setlength{\tabcolsep}{5pt}
    % \small
    \begin{tabular}{lcccccc}
    \toprule
    $p^{\mathcal{T}}$ & $p^{\mathcal{S}}$ & Ratio(\%) & Many & Med & Few & All \\
    \midrule
    - & - & - & 74.3 & 59.1 & 36.6 & 57.7 \\
    All & All & 100 & 74.8 & 59.8 & 38.4 & 58.6 \\
    Correct & - & 73 &74.6 & 59.8&  37.9 & 58.4\\
    Wrong & Wrong & 25 & 73.6 & 59.3 & 38.4 & 58.0\\
    Wrong & Correct & 2 &67.7 & 55.1&  34.2 & 53.2\\
    \midrule
    Ours & Ours & 98 & 75.5 & 60.1 &  38.7 & 59.1 \\
    \bottomrule
    \end{tabular}
    \caption{We compare the results of several Conflicting Knowledge Filters on the CIFAR-100-LT (IF=100). The baseline is VICAL w/o DED. The Ratio denotes the percentage of samples used for DED throughout the training. }
    \label{tab:ablation_ckf}
\end{table}

\begin{table*}[t]
    \setlength{\tabcolsep}{2pt}
    \centering
    \begin{tabular}{l | c  c c | c c c |c c c| c c c}
    \toprule
    \multirow{2} {*}{Method} & \multicolumn{3}{c|}{All} & \multicolumn{3}{c|}{Many} & \multicolumn{3}{c|}{Med} & \multicolumn{3}{c}{Few} \\
     & Acc. & Bias&  Var & Acc. & Bias&  Var &Acc. & Bias&  Var &Acc. & Bias&  Var \\
    \midrule
         CE & 31.6 & 0.60& 0.47& 57.3& 0.28& 0.35& 28.2& 0.61& 0.51 &6.3& 0.94& 0.57 \\ 
         %LDAM & 34.4 & 0.53& 0.51& 55.1 &0.28 &0.38 &31.9& 0.53 &0.54 &13.9& 0.81 &0.63 \\
         RIDE\cite{ride} & 40.5 &0.50& 0.42 &60.5 &0.28 &0.30 &38.7& 0.50 &0.44& 20.1 &0.74& 0.52 \\
         MDCS\cite{mdcs} & 46.1 & - & 0.36 & - & -& 0.24 & - & - & 0.38 & - & -& 0.46\\
         \midrule
         MDCS$^{\dagger}$& 48.2 & 0.43 & 0.37& 68.9 & 0.22& 0.24 & 55.1 & 0.34 & 0.34 & 30.4 & 0.63& 0.48\\
    % \midrule
    $\lambda$=2 & 49.9 & 0.42 & 0.35 & \textbf{74.6} & \textbf{0.16}& \textbf{0.19} & 57.0 & 0.33 & 0.31 & 29.3 & 0.65& 0.48\\
     $\lambda$=0 & 49.9 & 0.42 & 0.35 & 74.4 & 0.17& 0.20 & 56.8 & 0.32 & 0.32 & 29.5 & 0.65& 0.48\\
     $+$ Mixup\cite{mixup} & 49.7 & 0.41 & 0.36 & 69.4 & 0.22& 0.23 & 56.1 & 0.34 & 0.32 & 32.7 & \textbf{0.59}& 0.47\\
     $+$ AugMix\cite{augmix} & 49.3 & 0.42 & 0.36 & 71.7 & 0.19& 0.22 & 55.7 & 0.33 & 0.32 & 30.7 & 0.62& 0.48\\
     
     \midrule
    \rowcolor{yellow!30} \textbf{Ours} & \textbf{52.3} & \textbf{0.39} & \textbf{0.33} & 73.6 & 0.18& \textbf{0.19} & \textbf{59.3} & \textbf{0.31} & \textbf{0.29} & \textbf{33.9} & \textbf{0.59} & \textbf{0.45}\\
    \bottomrule
\end{tabular}\caption{Comparisons of the mean accuracy, bias, and variance of baselines and our proposed method on CIFAR-100-LT (IF=100), following the previous setup~\cite{ride,mdcs}. $\dagger$ represents our reproduced results using the official code on the specified datasets for a fair comparison. $\lambda=0$ refers to our proposed method without SC and DED.
}
 \label{tab:ablation_variance}
\end{table*}

\subsection{Further Analysis}
\noindent \textbf{Frequency-based Analysis} 
Fig.~\ref{fig: ablation_res} shows the results when using views at different resolutions for DED. 
Tab.~\ref{tab:ablation_resolution_freq} further shows the results of applying different filters, which confirms that discarding unstable high-frequency patterns effectively mitigates cross-expert knowledge conflicts.
To validate our frequency-decoupled mechanism,
Fig.~\ref{fig: fourier} visualizes the model's error rate on the CIFAR-100-LT validation set against Fourier basis noise, following~\cite{fourier}.
VICAL significantly expands the low-frequency robustness region for tail classes, while increasing sensitivity to extreme high-frequency noise for head classes. This asymmetric shift demonstrates that VICAL restricts the exploitation of unstable high-frequency shortcuts, compelling the model to establish robust dependencies on low-frequency structures.

\noindent \textbf{Identifying Conflicting Knowledge}
We provide Tab.~\ref{tab:ablation_ckf} to identify which teacher-student prediction pairs create conflicts. Specifically, we divide the samples into several groups based on whether the teachers' and students' predictions are correct. 
Although samples where the teacher is incorrect and the student is correct account for only 2$\%$, they can cause a significant performance drop and must be removed.
In Fig.~\ref{fig: conflict_cls_analysis}, we provide the histogram of the conflicting knowledge set $\mathbb{D}_c$. During the terminal training phase, for misclassified samples from difficult head classes, the ensemble consensus tends to predict them as another head class.
CKF safeguards valuable individual expert knowledge from being compromised by uncertain consensus.

\begin{figure}[t]
    \centering
    \includegraphics[width=0.865\linewidth]{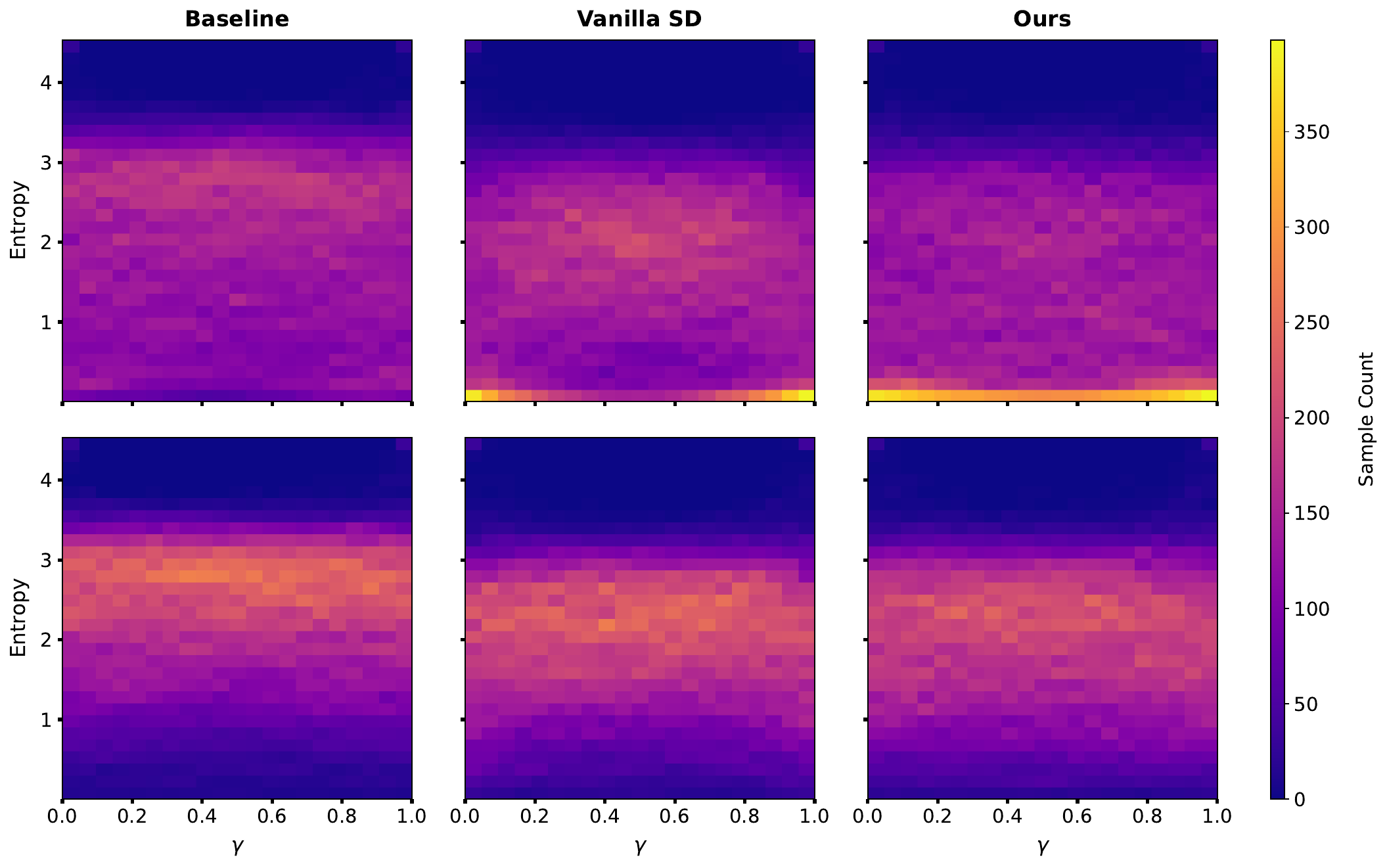}
    \caption{Heatmaps of the entropy profiles as a function of the interpolation factor $\gamma$ $\in$ [0,1] between two full-resolution views $\mathrm{v}_{1}$ and $\mathrm{v}_{2}$ on the CIFAR-100-LT test set with a single expert. Head samples (\textbf{Top}) are more susceptible to $\gamma$, whereas tail samples (\textbf{Bottom}) have higher entropy overall.
    SC reduces the entropy of interpolated samples, particularly for tail classes.}
    \label{fig:entropy_heatmap}
\end{figure}
\noindent \textbf{Heatmaps of Entropy Profiles}
To validate the efficacy of SC in improving consistency and reducing prediction uncertainty, we present heatmaps to show how the entropy of interpolated samples on the test dataset varies with respect to $\gamma$ in Fig.~\ref{fig:entropy_heatmap}. 
Specifically, we train three ResNet-32 networks independently using Balanced Softmax\cite{b_softmax} (BS) loss, BS jointly with vanilla SD, and BS jointly with our SC module on CIFAR-100-LT.
The entropy of tail samples on the test set is typically higher than that of head samples.
SC significantly reduces the uncertainty and induces a smoother entropy profile, especially for tail classes.

\noindent \textbf{Model Bias and Variance}
We aim to enhance model robustness and reduce variance through the proposed VICAL. To validate the effectiveness, we follow the prior setup~\cite{ride,mdcs} and train 20 independent models on subsets of CIFAR-100-LT.
Tab.~\ref{tab:ablation_variance} shows the comparison results. Increasing $\lambda$ enhances output diversity as shown in Fig.~\ref{fig:div-acc}, yet yields no corresponding improvement in model bias or variance. 
Mixup~\cite{mixup} and AugMix~\cite{augmix} enhance model robustness by interpolating inputs across different classes and combining multiple random perturbations into a single input, respectively.
Although Mixup and AugMix effectively mitigate bias in tail classes, this advantage comes at the cost of significant performance degradation in the head.
In contrast, VICAL not only reduces variance and bias on tail classes but also preserves the head's performance, demonstrating its superior regularization effect on long-tailed learning.

\section{Conclusion}
In this paper, we critically re-evaluate the prevailing hypothesis in long-tailed learning that assumes a positive correlation between model diversity and ensemble accuracy. We unveil that diversity induced by logit adjustment or explicit regularizers does not universally guarantee better ensemble accuracy.
Motivated by these findings, we introduce Vicinal Consistency Alignment (VICAL), a novel framework that shifts the focus from explicit diversity constraints to implicit variance reduction. 
By penalizing reliance on unstable high-frequency patterns and enforcing low-frequency semantic agreement, VICAL improves the stability within each expert and sidesteps optimization conflicts through views at asymmetric resolutions.
Extensive experiments show that VICAL consistently achieves state-of-the-art performance across benchmarks. More importantly, it effectively reduces both variance and bias, particularly for tail classes. Our work demonstrates that vicinal consistency alignment offers a powerful alternative for improving generalization in long-tailed recognition.

\section*{Acknowledgements}
This project was supported by the National Natural Science Foundation of China (NSFC) Projects under Grants No. 62522206 and No. 62521004.
% Please insert your acknowledgments here.


\newpage

\title{VICAL: Vicinal Consistency Alignment for Long-Tailed Visual Recognition}
\author{Supplementary Material}
\institute{
}
\date{}
\titlerunning{VICAL}
\authorrunning{Supplementary Material}

% \begin{document}
\maketitle
\section{Datasets and Training Details}
\subsection{Datasets}
\noindent \textbf{Long-tailed CIFAR} 
We conduct experiments on CIFAR-10-LT and CIFAR-100-LT~\cite{ldam}, the long-tailed variants of CIFAR-10 and CIFAR-100~\cite{cifar}, containing 10 and 100 classes, respectively. Following common practice~\cite{bbn, ldam, cbloss}, we adopt standard long-tailed splits for fair comparison. The imbalance factor is $N_{max}/N_{min}$ and reflects the degree of imbalance in the data. The imbalance factors used in our experiments are 200, 100, 50, and 10.

\noindent\textbf{ImageNet-LT} ImageNet-LT~\cite{oltr} is a long-tailed version of the original ImageNet dataset, constructed by sampling a subset according to a Pareto distribution with power value $\alpha =6$. The resulting dataset contains 115.8K images from 1000 categories, with 1280 to 5 images per class.

\noindent\textbf{iNaturalist 2018} The iNaturalist 2018 dataset~\cite{inaturalist} is a large-scale, real-world fine-grained benchmark with a natural long-tailed distribution. It consists of 437.5K images across 8,142 categories, exhibiting a high imbalance factor of 500. This dataset presents dual challenges of severe class imbalance and fine-grained visual recognition.

\subsection{Training Details}
\noindent \textbf{Backbone} 
For CIFAR-LT, we follow~\cite{balpoe} and use ResNet-32~\cite{resnet} as the backbone, which shares the first block across experts and does not reduce the expert's channel dimension. We find it is necessary to achieve high performance on CIFAR-LT, and we reproduce BalPoE~\cite{balpoe} and MDCS~\cite{mdcs} using the same backbone for fair comparison.
For ImageNet-LT and iNaturalist 2018, we utilize ResNeXt-50~\cite{resnext} and ResNet-50 as the backbones. The first two shallow blocks serve as a shared net, and the deeper blocks form the expert-specific branches. The channel dimension is reduced by $1/4$, following common practice~\cite{ride,sade,mdcs,mutual}.
Note that BalPoE does not adopt the channel-reduction architecture and consequently has more trainable parameters (See Tab.~\ref{tab:img_inat_paramter}). 
We implement our method on CIFAR-LT, ImageNet-LT, and iNaturalist 2018 using 1, 4, and 8 NVIDIA RTX 3090 GPUs, respectively.

\noindent \textbf{Data Augmentation} 
For data augmentation, we mainly follow~\cite{paco,bcl,mdcs} to implement strong augmentation, \textit{i.e.}, AutoAugment~\cite{autoaugment} for CIFAR-LT and RandAugment~\cite{randaug} for large-scale datasets. The views $\mathrm{v_{1}}$ and $\mathrm{v_{2}}$ are generated by independently applying strong augmentation twice.
The low-resolution view undergoes the same augmentation pipeline, varying only in spatial resolution, 16$\times$16 for CIFAR-LT and 96$\times$96 for large-scale datasets. 

\begin{figure*}
    \centering
    \includegraphics[width=1.0\linewidth]{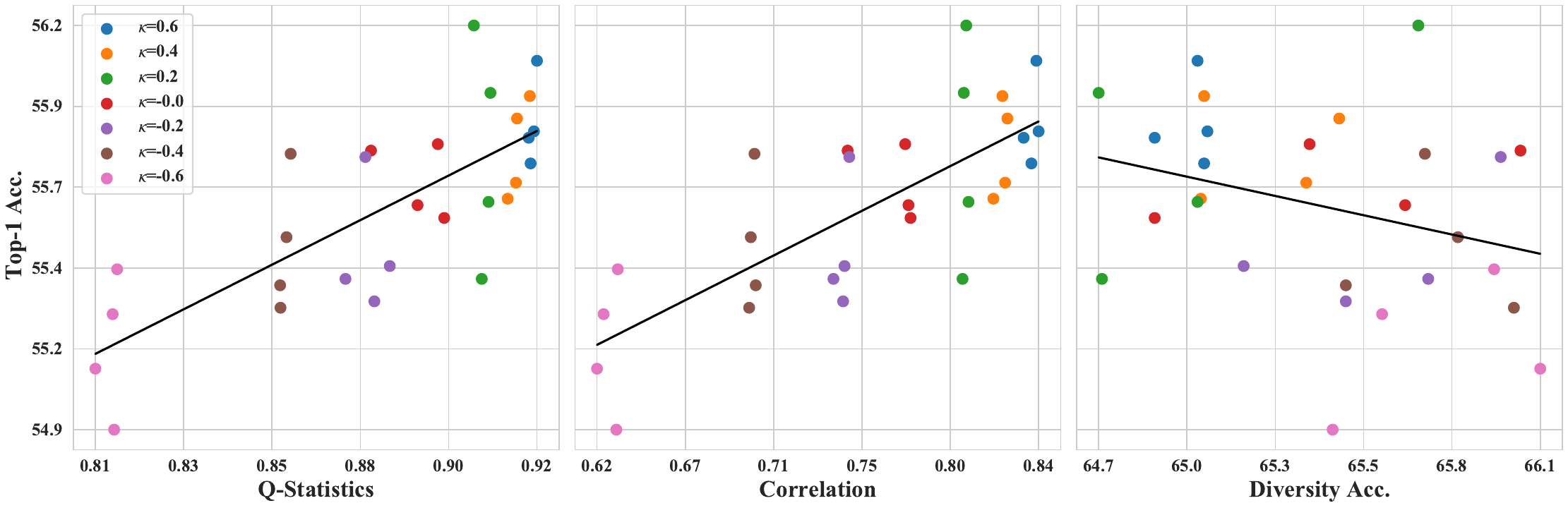}
    \caption{We use three different diversity metrics, \textit{i.e.}, Q-statistics, Correlation, and Diversity accuracy, to measure the relationship between model diversity and ensemble accuracy on CIFAR-100-LT with an \textbf{explicit diversity regularizer}. Each point represents an independently trained multi-expert model.
    $\kappa$ controls the explicit diversity regularizer intensities.
    A larger $\kappa$ value encourages each expert to align with the ensemble prediction $\overline{p}$, while a negative $\kappa$ forces predictions to diverge from this ensemble consensus. Ensemble accuracy shows a positive correlation with both Q-statistics and the correlation coefficient $\rho$. (The Pearson correlation coefficients are $0.76$, $0.76$, and $-0.31$ for Q-statistics, Correlation, and Diversity accuracy.)}
    \label{fig:div-acc2}

\end{figure*}

\begin{figure}[t]
    \centering
    \subfloat[]{
\includegraphics[width=0.395\linewidth]{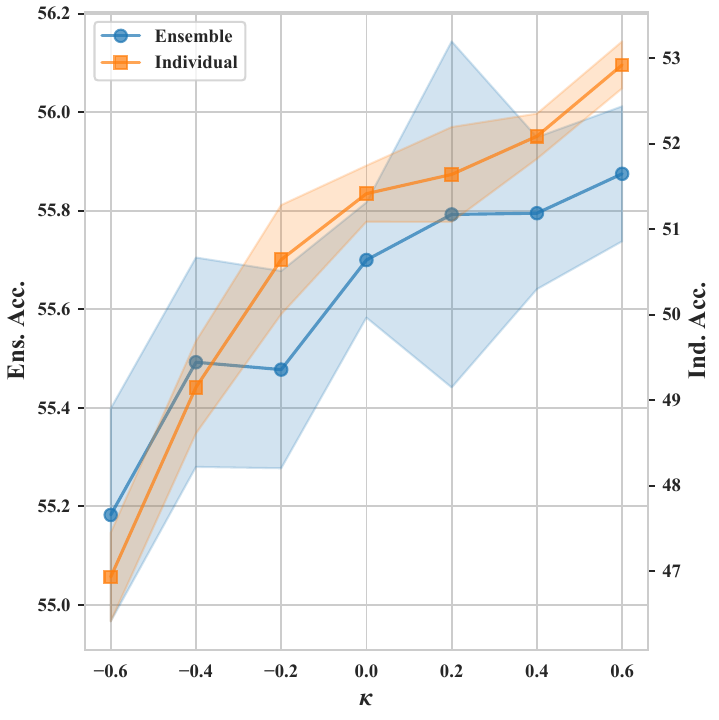} \label{fig:ens_vs_ind}  } 
    \subfloat[]{
\includegraphics[width=0.385\linewidth]{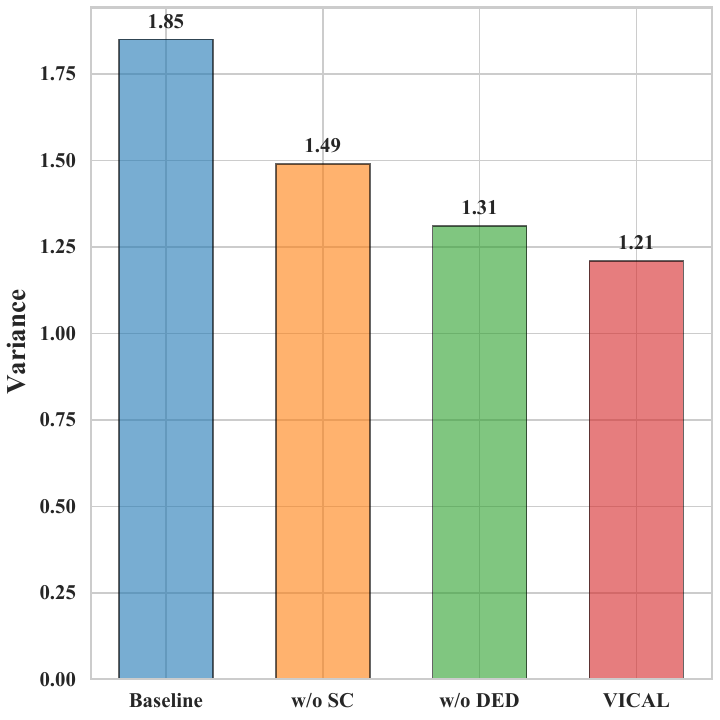} \label{fig: variance_reduction}
    }
    \caption{(a) Ensemble accuracy and mean individual expert accuracy across different $\kappa$. A large positive $\kappa$ yields better individual accuracy, thereby leading to an improved ensemble performance. (b) Inter-expert logit variance on the CIFAR-100-LT test set. We compute the variance of logits across experts. Lower values indicate stronger cross-expert consistency.
    }
    
\end{figure}

\section{Experimental Setup of Section 3}
\subsection{Diversity Measures}
In this paper, we employ two commonly used pairwise metrics, \textit{i.e.}, \textit{Q-statistics}~\cite{q-statistics} and \textit{correlation coefficient $\rho$}, to measure the diversity of multi-expert models.

\noindent \textbf{Q-statistic} 
The Q-statistic quantifies the correlation between two classifiers in terms of classification predictions, ranging from $[-1, 1]$.
Let $N^{11}$ ($N^{00}$) represent the number of samples that can be recognized by both classifiers correctly (incorrectly).
Similarly, $N^{10}$ and $N^{01}$ are defined as the number of samples recognized correctly by only one of the classifiers and incorrectly by the other classifier.
The Q-statistic value of classifiers $i$ and $j$ can be calculated as follows:
\begin{equation}
    Q_{ij} = \frac{N^{11}N^{00}-N^{01}N^{10}}{N^{11}N^{00}+N^{01}N^{10}}.
\end{equation}
The overall Q-statistic is the average of the Q-statistic value over all pairs of classifiers, \textit{i.e.}, 
\begin{equation}
    Q=\frac{1}{K(K-1)}\sum^{K}_{i}\sum^{K}_{j\neq i}Q_{ij},
\end{equation}
where $K$ is the number of classifiers and equals the number of experts $M$ in our setting.
For independent classifiers, the expectation of $Q$ is $0$. A positive Q-statistic indicates that the classifiers tend to be correct or incorrect on the same samples, whereas a negative value indicates that one classifier tends to be correct when the other is incorrect~\cite{ensemble_metrics}.

\noindent \textbf{Correlation Coefficient} 
We also use the Pearson correlation coefficient $\rho$ to quantify the dependency of two classifiers $i$ and $j$.
Specifically, for each class $k\in\{1,…,C\}$, we calculate the Pearson correlation between the two classifiers' predicted probability vectors across all test samples. The $\rho_{ij}$ is obtained by averaging these class-wise coefficients of classifiers $i$ and $j$,
\begin{equation}
    \rho_{ij} = \frac{1}{C}\sum_{k=1}^{C}Pearson(P^{i}_{:,k},P^{j}_{:,k}),
\end{equation}
where $P^{i}$ and $P^{j}$ are probability matrices of two classifiers $i$ and $j$, respectively.
The overall correlation coefficient $\rho$ is averaged over all classifier pairs.

\begin{figure}[htbp] 
    \centering
    \subfloat[]{\includegraphics[scale=0.25]{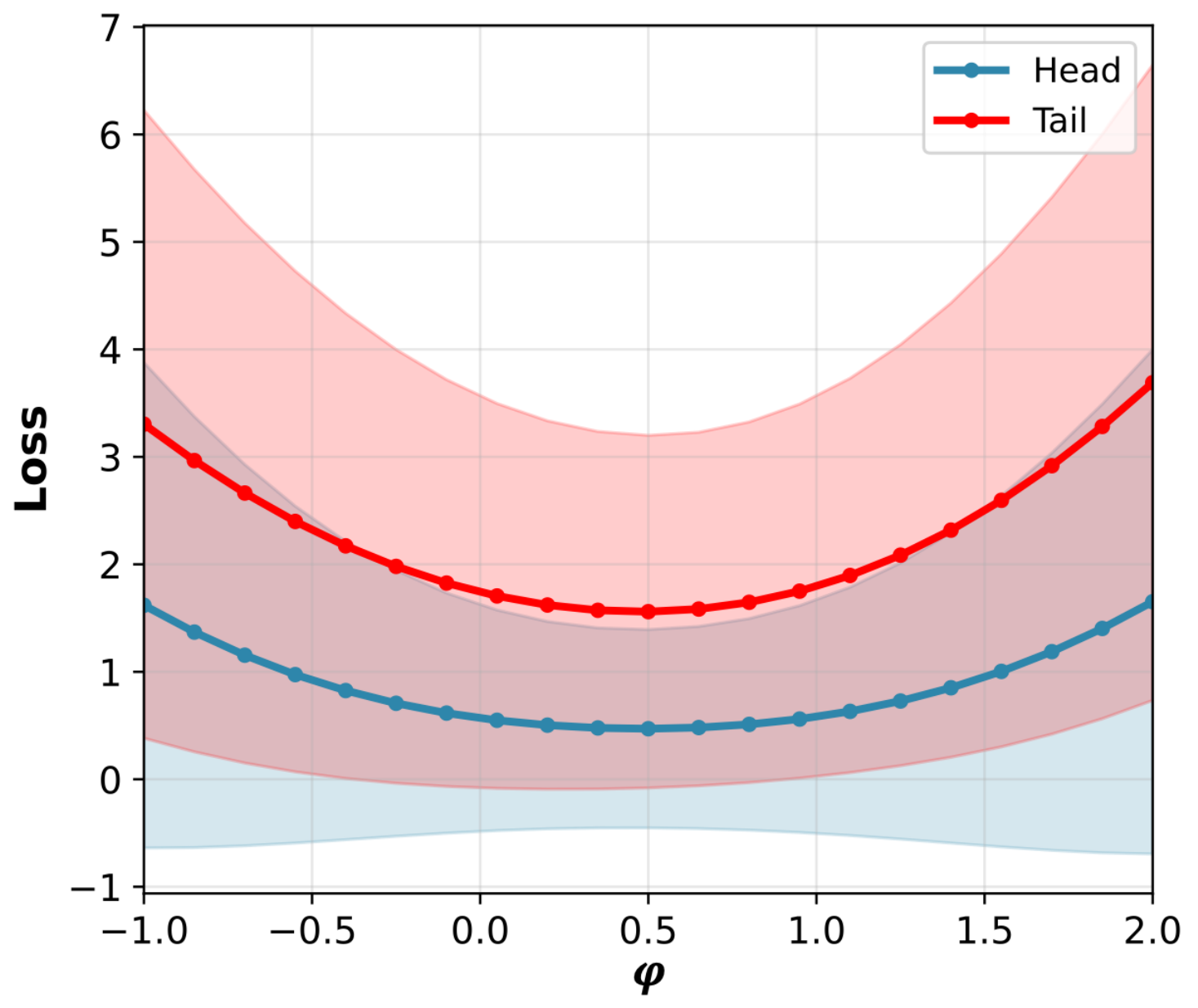} \label{fig: ablation_phi} }
     \subfloat[]{\includegraphics[scale=0.55]{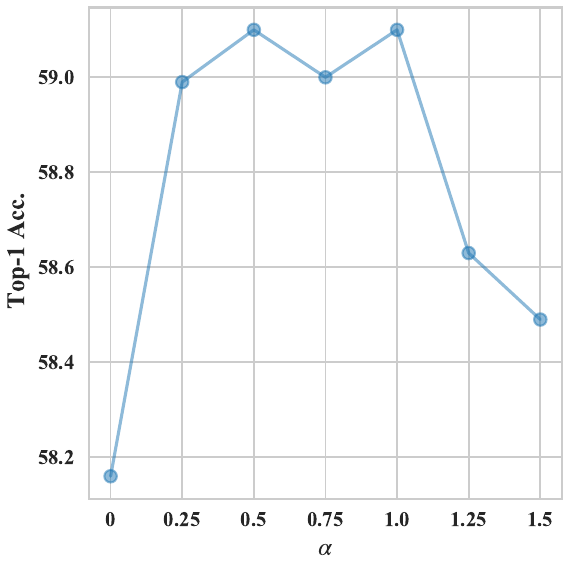} \label{fig:supp_beta_distribution}}
    \caption{
    Ablations of (a) the teacher logit aggregation weight $\varphi$ and (b) the Beta-distribution parameter $\alpha$ on CIFAR-100-LT (IF=100).}
    \label{fig: ablation_others}
\end{figure}

\noindent \textbf{Diversity Factor} 
In addition to pairwise measures, we also employ the diversity factor~\cite{mdcs} as an auxiliary metric, which is a kind of accuracy and is calculated as follows:
\begin{equation}
\sigma=\frac{\left|\bigcup_{m=1}^{M}S_m\right|}{|\mathbb D_u|}, 
\quad
    S_m=\left\{i\mid \arg\max_{c}p_c^m(x_i)=y_i,
(x_i,y_i)\in\mathbb D_u\right\},
\end{equation}
where $S_{m}$ is the set of all correctly recognized samples in the uniform test set $\mathbb{D}_{u}$ by the $m$-th expert.
The diversity factor $\sigma$ equals the top-1 accuracy when all experts' outputs are identical, and attains its maximum when every sample is correctly classified by at least one expert.

\subsection{Explicit Diversity Regularizer}
In addition to varying logit adjustment intensities, we also provide a standard explicit constraint used in RIDE~\cite{ride} to enforce diversity among experts directly. Let $p^{m}$ denote the $m$-th expert's prediction probability. The diversity can be achieved by
\begin{equation}
    \mathcal{L}_{div}=\kappa \sum_{m=1}^{M} KL(\overline{p}||p^{m}),
\end{equation}
where $\overline{p}$ refers to the ensemble prediction probability and $\kappa$ is the hyperparameter that controls explicit diversity intensity.
A large positive $\kappa$ encourages each expert to align with the ensemble consensus $\overline{p}$, while a negative $\kappa$ enforces divergence from this ensemble prediction.
We jointly optimize the model using the diversity loss alongside Balanced Softmax loss~\cite{b_softmax} for each expert.

RIDE~\cite{ride} aims to achieve better performance via a negative $\kappa$; however, our experiments unveil an unexpected result.
As shown in Fig.~\ref{fig:div-acc2}, a negative value of $\kappa$ leads to lower Q-statistics and correlation coefficient $\rho$, indicating increased prediction diversity but at the cost of degraded ensemble performance.
Furthermore, ensemble accuracy exhibits a slight negative correlation with diversity accuracy $\sigma$.
We hypothesize that enforcing diversity via an explicit regularizer inadvertently damages the learned representation quality of each expert.
Conversely, positive $\kappa$s enable each expert to reduce diversity while benefiting from ensemble consensus, thereby improving overall performance.
Fig.~\ref{fig:ens_vs_ind} confirms our hypothesis: a large positive $\kappa$ yields superior individual experts. At $\kappa=-0.6$, ensemble accuracy incurs only marginal degradation while individual expert performance suffers substantial deterioration.
In summary, explicit diversity enforcement proves suboptimal, motivating our shift to a consistency alignment framework.

\section{Extended Experiments}
\noindent \textbf{Inter-Expert Variance Analysis}
We directly measure the variance among expert logits on CIFAR-100-LT, as shown in Fig.~\ref{fig: variance_reduction}. The baseline exhibits the highest variance of 1.85. Introducing DED and SC reduces it to 1.49 and 1.31, respectively. The complete VICAL framework achieves the lowest variance
of 1.21.
These results confirm that both components improve cross-expert consistency and jointly provide the strongest variance reduction.

\noindent \textbf{Effect of SC Hyperparameters}
We first vary the aggregation weight in $\bar z^m=\varphi z_{v_1}^m+(1-\varphi)z_{v_2}^m$ to form the mixed teacher target.
As shown in Fig.~\ref{fig: ablation_phi}, equal averaging (\(\varphi=0.5\)) minimizes the loss for both head and tail classes, whereas extrapolative weights outside \([0,1]\) increase the loss. 
Fig.~\ref{fig:supp_beta_distribution} shows that performance is stable for $\alpha\in[0.25,1.0]$. We use $\alpha=1.0$ in all experiments.

\noindent \textbf{Effect of SC on Individual Experts} 
We further validate our proposed Self-Consistency Learning module on a single expert.  We compare several self-distillation variants using the same backbone and strong data augmentation strategy.
As listed in Tab.~\ref{tab:ablation_sd}, our proposed SC achieves the best results among these consistency alignment approaches.

\begin{figure}[htbp]
    \centering  \includegraphics[width=0.625\linewidth]{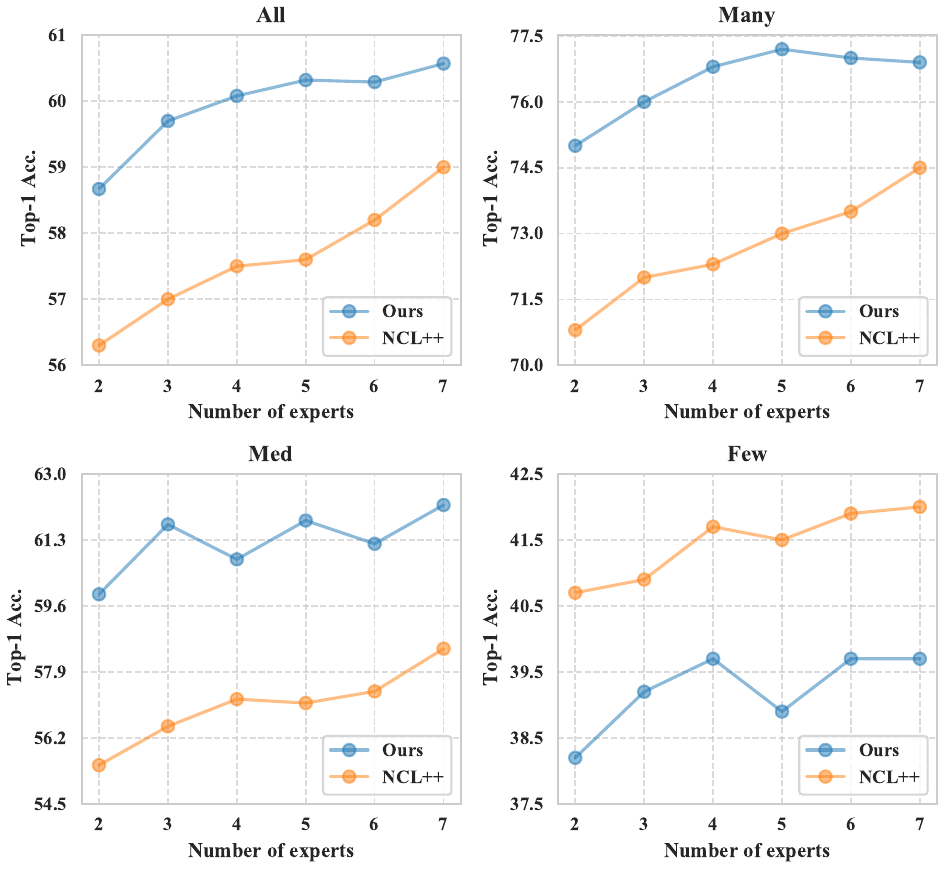}
    \caption{Comparisons of our method and NCL++ on CIFAR-100-LT (IF=100) with different numbers of experts. While NCL++ achieves superior accuracy in the tail classes, this is accompanied by a high cost in many and medium-shot performance. 
    }
    \label{fig:num_experts}
\end{figure}

\noindent \textbf{Effect of the Number of Experts} 
The VICAL framework demonstrates inherent scalability without additional hyperparameter tuning.
Fig.~\ref{fig:num_experts} illustrates the 
performance of VICAL compared to NCL++~\cite{ncl++} on CIFAR-100-LT (IF=100) under varying numbers of experts. 
Our framework consistently outperforms NCL++ across all expert configurations and exhibits superior prediction stability in the head and medium classes.

\noindent \textbf{Computation Cost} 
The VICAL framework is a multi-expert model with multiple views as input, which inevitably incurs additional computation cost.
For a fair comparison, we assess the performance of VICAL against NCL++~\cite{ncl++} and MDCS~\cite{mdcs} at an equivalent computation cost.
Specifically, we quantify the forward FLOPs for the online network per sample with multiple views as input.
Tab.~\ref{tab:ablation_cifar_flops} demonstrates that VICAL consistently surpasses MDCS and NCL++ under equivalent computation cost, demonstrating the efficacy of our consistency-driven approach.
We further compare model parameters and corresponding performance on ImageNet-LT.
As shown in Tab.~\ref{tab:img_inat_paramter}, 
BalPoE~\cite{balpoe} and NCL++~\cite{ncl++} partly benefit from having more trainable parameters, whereas our method achieves significant gains under comparable model capacity.

\noindent \textbf{Results on Multiple Test Distributions} 
Following the previous setup~\cite{sade,balpoe}, we compare model generalization under multiple test distributions.
Tab.~\ref{tab:cifar-lt_test-agnostic} and Tab.~\ref{tab:img-lt_test-agnostic} report the compared results on CIFAR-100-LT and ImageNet-LT, respectively.
Our proposed VICAL surpasses other methods across different test distributions by a large margin, demonstrating the benefits of vicinal consistency alignment.

\begin{table}[]
    \begin{minipage}{0.6\textwidth}
    \centering
    \small
    \begin{tabular}{lcccccc}
    \toprule
    Method & Many  & Med & Few & All & MFLOPs$\times$Epochs\\
    \midrule
    \rowcolor{yellow!30}MDCS & - & - & -& 56.1 & 204$\times$400 \\ 
    \rowcolor{green!20}MDCS$^{\dagger}$ & 74.8 & 59.0 & 34.5& 57.1 & 321$\times$400 \\ 
    \rowcolor{blue!20}NCL++ & 70.8 & 55.5 & \textbf{40.7} &
    56.3 &  561$\times$400 \\
    \midrule
    \rowcolor{yellow!30}Ours & 73.4 & 59.2 & 37.4 & 57.6 &  522$\times$155 \\
    \rowcolor{green!20} Ours & 75.5 & 60.1 & 38.7 & 59.1 &  522$\times$250 \\
    \rowcolor{blue!20}Ours & \textbf{76.0} & \textbf{61.7} & 39.2 & \textbf{59.7} &  522$\times$400 \\
    \bottomrule
    \end{tabular}
    \caption{Computation cost on CIFAR-100-LT (IF=100). $\dagger$ refers to our reproduced results.
    We compare the performance of our method against MDCS and NCL++ at an equivalent computation cost. The same color indicates comparable computation cost.}
    \label{tab:ablation_cifar_flops}
% \end{table}

    \end{minipage}
    \hfill
    \begin{minipage}{0.35\textwidth}
    \centering
    \begin{tabular}{lc}
    \toprule
    Methods & Top-1 Acc. \\
    \midrule
    BS & 50.6 \\
    Vanilla SD & 51.7 \\
    MDCS$^{\dagger}$\cite{mdcs} & 51.6 \\
    NCL++$^{\dagger}$\cite{ncl++} & 53.1 \\
    \midrule
    SC & 53.6 \\ 
    \bottomrule
    \end{tabular}
    \caption{Ablation study of variants of self-distillation with a single expert on CIFAR-100-LT training for 200 epochs. BS: Balanced Softmax. $\dagger$ refers to our reproduced results using a single expert with intra-expert distillation.}
    \label{tab:ablation_sd}
    \end{minipage}
\end{table}

\begin{figure}[htbp] 
    \centering
     \includegraphics[scale=0.375]{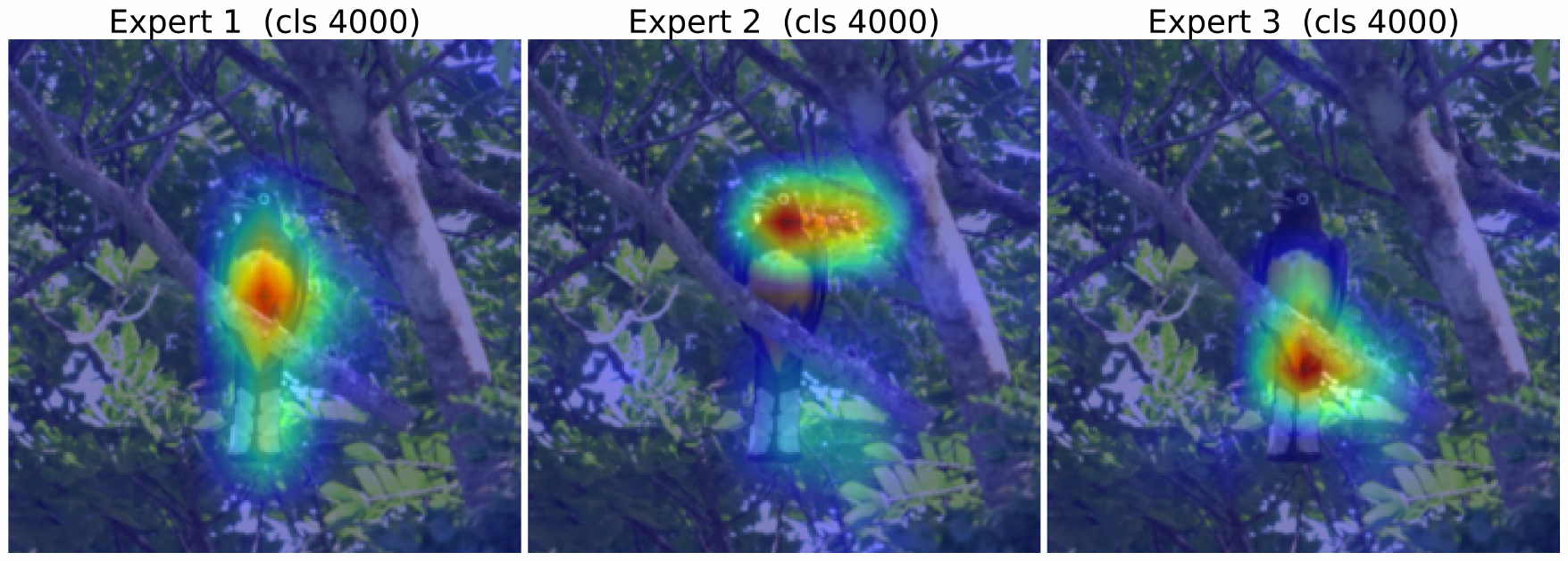}
    \includegraphics[scale=0.375]{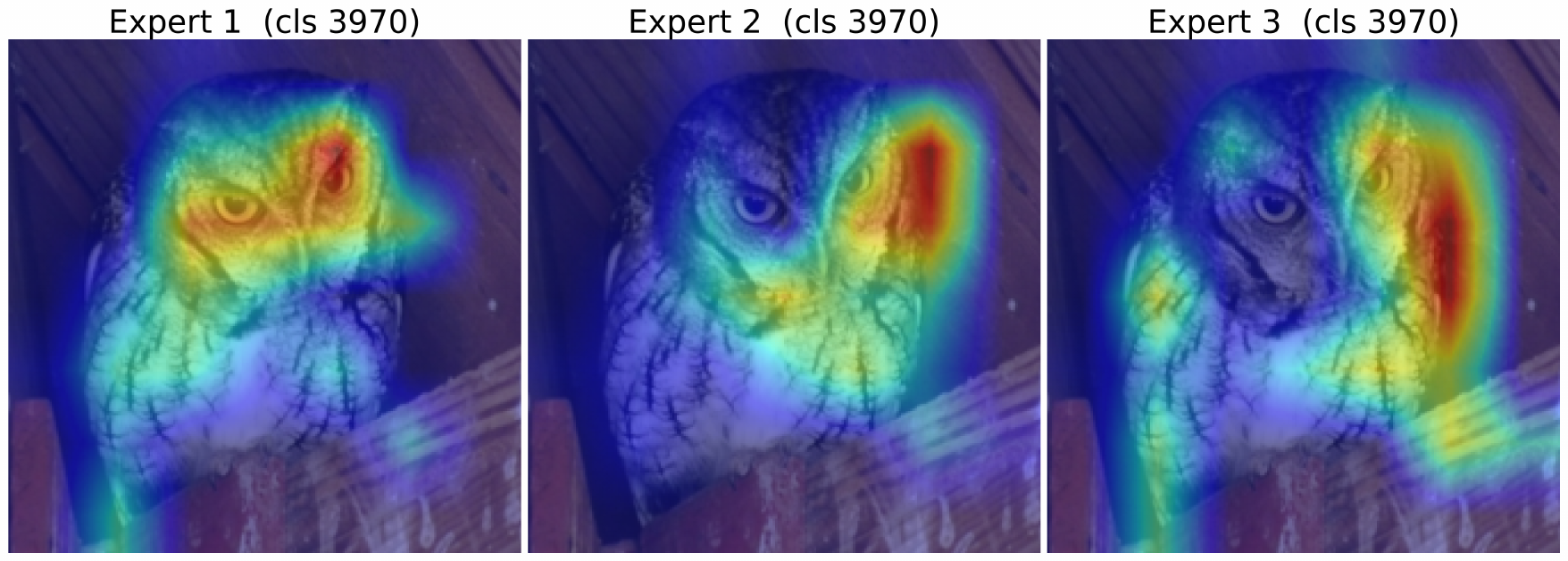}
     \caption{Grad-CAM visualizations for three experts on iNaturalist 2018.}
    \label{fig:gradcam}
\end{figure}

\begin{table}[htbp]
    \centering
    \begin{tabular}{l|c|ccc|c}
    \toprule
      Method & Params & Many & Med & Few & All\\

    \midrule
    
    BalPoE\cite{balpoe}& 66.1M & 68.2 & 57.2 & \textbf{44.9} & 59.8 \\ 
    BalPoE$^{\ddag}$\cite{balpoe}& 66.1M & - & - & - & 61.6 \\
    
    NCL++$^{\ddag}$\cite{ncl++}& 50.1M & - & - & - & 60.9 \\
    RIDE\cite{ride} & 38.3M & 67.6 & 53.5 & 35.9 & 56.4\\
    SADE\cite{sade} & 38.3M & 66.5 & 57.0 & 43.5 & 58.8\\
    ML\cite{mutual} & 38.3M & 70.2 & 56.7 & 39.1 & 59.5\\
    MDCS$^{\ddag}$\cite{mdcs}& 38.3M & - & - & - & 60.2 \\
    \midrule
    \rowcolor{yellow!30}
    \textbf{Ours}$^{\ddag}$ &  38.3M&\textbf{72.8} &  \textbf{60.8} & 42.6 & \textbf{62.9}   \\ 
    \bottomrule
    \end{tabular}
    \caption{Top-1 accuracy on ImageNet-LT using ResNeXt-50 as backbone. We report the results of 200 epochs and the trainable parameters of all compared multi-expert models. $\ddag$ denotes models trained with RandAugment\cite{randaug}.}
    \label{tab:img_inat_paramter}
\end{table}

\noindent \textbf{Visualization Results}
VICAL processes both full and low-resolution views: the low-resolution view ensures semantic consistency (consistent label), while the full-resolution view preserves the experts' ability to capture diverse details.
We use Grad-CAM~\cite{gradcam}
to visualize each expert's capability of capturing different patterns, as shown in Fig.~\ref{fig:gradcam}.

\begin{table}[t]
    \centering
     \small
    \begin{tabular}{lcccccccccccc}
     \toprule
   \multirow{2}{*}{Method}   &    \multirow{2}{*}{Prior}   & \multicolumn{5}{c}{Forward-LT} & \multicolumn{1}{c}{Uni.} & \multicolumn{5}{c}{Backward-LT} \\
\cmidrule(lr){3-7} \cmidrule(lr){8-8} \cmidrule(lr){9-13}
   &  & {50}   & {25}   & {10}   & {5}    & {2}    & {1}    & {2}    & {5}    & {10}   & {25}   & {50}   \\
\midrule
     
     LADE & $\times$ &  56.0 & 55.5 & 52.8 & 51.0 & 48.0 & 45.6 & 43.2 & 40.0 & 38.3 & 35.5 & 34.0 \\
     RIDE & $\times$ & 63.0 & 59.9 & 57.0 & 53.6 & 49.4 & 48.0 & 42.5 & 38.1 & 35.4 & 31.6 & 29.2 \\
     SADE & $\times$ & 58.4 & 57.0 & 54.4 & 53.1 & 50.1 & 49.4 & 45.2 & 42.6 & 39.7 & 36.7 & 35.0 \\
     BalPoE & $\times$ & 65.1 &63.1 &60.8 &58.4 &54.8 & 52.0 & 48.6 & 44.6&41.8& 38.0 &36.1 \\
     BalPoE $^{\dag}$ & $\times$ & 
    71.4& 69.4&66.1&63.6&59.7&56.4&53.0&
    49.0&46.2&41.9&39.5
     \\
     MDCS$^{\dag}$ & $\times$    &71.8&69.5&66.7&64.4&60.2&57.1&53.4&49.3&46.2&41.8&40.3
     \\
     \rowcolor{yellow!30}
     \textbf{Ours} & $\times$ & \textbf{73.3} & \textbf{71.6} & \textbf{69.1} & \textbf{66.4} & \textbf{62.5} & \textbf{59.7} & \textbf{56.2} & \textbf{52.6} & \textbf{49.7} & \textbf{45.4} & \textbf{43.6} \\
     \midrule
     LADE & $\checkmark$ & 62.6& 60.2& 55.6& 52.7&48.2& 45.6 & 43.8 & 41.1 &41.5 &40.7 &41.6 \\
     BalPoE & $\checkmark$ & 70.3& 66.8 &62.7 &59.3&54.8&52.0&49.2 & 46.9 &46.2 &45.4 &46.1\\
     BalPoE $^{\dag}$ & $\checkmark$ &
    74.1&71.1&67.1&64.0&59.5&56.4&53.7&
    51.4&50.7&49.7&50.8
     \\
      MDCS$^{\dag}$ & $\checkmark$   
    &75.3&72.1&68.0&64.8&59.8&57.1&54.4&52.1& 51.2&50.7&51.4\\
    \rowcolor{yellow!30}
     \textbf{Ours} & $\checkmark$ & \textbf{75.8} & \textbf{72.7} & \textbf{69.3} & \textbf{66.0} & \textbf{62.3} & \textbf{59.7} & \textbf{56.8} & \textbf{55.4} & \textbf{54.7} & \textbf{54.3} & \textbf{55.6} 
     \\ 
     \bottomrule
\end{tabular}
\caption{Top-1 accuracy on multiple test distributions for ResNet-32 trained on CIFAR-100-LT (IF=100).  Prior denotes whether the test set prior is used. $\dagger$ refers to our reproduced results, and other results are from~\cite{balpoe}.}
    \label{tab:cifar-lt_test-agnostic}
\end{table}

\begin{table}[t]
    \centering
    \small
    \begin{tabular}{lcccccccccccc}
     \toprule
   \multirow{2}{*}{Method}   &    \multirow{2}{*}{Prior}   & \multicolumn{5}{c}{Forward-LT} & \multicolumn{1}{c}{Uni.} & \multicolumn{5}{c}{Backward-LT} \\
\cmidrule(lr){3-7} \cmidrule(lr){8-8} \cmidrule(lr){9-13}
   &  & {50}   & {25}   & {10}   & {5}    & {2}    & {1}    & {2}    & {5}    & {10}   & {25}   & {50}   \\
\midrule
     
     LADE & $\times$ &63.4	&62.1&	59.9	&57.4	&54.6&	52.3	&49.9	&46.8	&44.9	&42.7	&40.7\\
     RIDE & $\times$ & 67.6&	66.3&	64.0&	61.7	&58.9&	56.3	&54.0&	51.0&	48.7	&46.2&	44.0 \\
     SADE & $\times$ & 65.5	&64.4&	63.6&	62.0	&60.0	&58.8&	56.8	&54.7	&53.1	&51.1	&49.8 \\
     BalPoE & $\times$ & 67.6&66.3&65.2&63.3&61.5&59.8&58.1&55.7&54.3&52.2&50.8 \\
     \rowcolor{yellow!30}
     \textbf{Ours} & $\times$ & 
     \textbf{72.1} & \textbf{70.9}& \textbf{69.3} & \textbf{67.3}& \textbf{65.0} & \textbf{62.9} &
     \textbf{60.7} & \textbf{57.7} & \textbf{55.8} & \textbf{53.3} &  \textbf{51.3}
     \\
     \midrule
     LADE & $\checkmark$ & 65.8&	63.8&	60.6&	57.5	&54.5	&52.3	&50.4	&48.8&	48.6&	49.0&	49.2 \\
     SADE & $*$ & 69.4&	67.4	&65.4&	63.0&	60.6&	58.8&	57.1	&55.5	&54.5&	53.7&	53.1     \\ 
     BalPoE & $\checkmark$ & 72.5&70.2&67.3&64.6&61.8&59.8&58.3&57.2&56.6&56.6&56.9\\
     \rowcolor{yellow!30}
     \textbf{Ours} & $\checkmark$ & \textbf{74.7}&\textbf{72.8}&\textbf{70.1} & \textbf{67.3} & \textbf{64.8} & \textbf{62.9} & \textbf{61.4} & \textbf{60.0} & \textbf{60.0} & \textbf{59.8} & \textbf{59.7}
     \\ 
     \bottomrule
\end{tabular}
\caption{Top-1 accuracy on multiple test distributions for ResNeXt-50 trained on ImageNet-LT. Prior denotes whether the test set prior is used, and $*$ refers to implicit test prior estimation. }
    \label{tab:img-lt_test-agnostic}
\end{table}

\clearpage

\section{Broader Impact and Limitations}
Our proposed VICAL demonstrates its superiority in imbalanced classification, suggesting its potential for fairness and reliability of vision systems in critical applications. VICAL can also be directly applied to semi-supervised visual recognition tasks without requiring complex modifications.
The framework’s emphasis on stability and variance reduction may also contribute to more robust and trustworthy AI systems in real-world deployments.

Despite remarkable efficacy, our VICAL framework has several limitations that suggest valuable future directions.
The multi-expert architecture with vicinal views increases training time compared to single-model baselines, although inference cost remains comparable.
Moreover, the precise theoretical relationship between consistency learning and variance reduction in long-tailed settings deserves further formal analysis.

\bibliographystyle{splncs04}
\bibliography{main}

@String(IJCV  = {Int. J. Comput. Vis.})

@String(ICLR  = {Int. Conf. Learn. Represent.})

@String(BMVC  = {Brit. Mach. Vis. Conf.})

@String(AAAI  = {AAAI})

@String(IJCV  = {IJCV})

@String(ICLR  = {ICLR})

@String(BMVC  =	{BMVC})

@article{q-statistics,
  title={VII. On the association of attributes in statistics: with illustrations from the material of the childhood society, \&c},
  author={Yule, George Udny},
  journal={Philosophical Transactions of the Royal Society of London. Series A, Containing Papers of a Mathematical or Physical Character},
  volume={194},
  number={252-261},
  pages={257--319},
  year={1900},
  publisher={The Royal Society London}
}

@article{ensemble_metrics,
  title={Measures of diversity in classifier ensembles and their relationship with the ensemble accuracy},
  author={Kuncheva, Ludmila I and Whitaker, Christopher J},
  journal={Machine learning},
  volume={51},
  pages={181--207},
  year={2003},
  publisher={Springer}
}

@inproceedings{mdcs,
  title={Mdcs: More diverse experts with consistency self-distillation for long-tailed recognition},
  author={Zhao, Qihao and Jiang, Chen and Hu, Wei and Zhang, Fan and Liu, Jun},
  booktitle={Proceedings of the IEEE/CVF International Conference on Computer Vision},
  pages={11597--11608},
  year={2023}
}

@inproceedings{balpoe,
  title={Balanced product of calibrated experts for long-tailed recognition},
  author={Aimar, Emanuel Sanchez and Jonnarth, Arvi and Felsberg, Michael and Kuhlmann, Marco},
  booktitle={Proceedings of the IEEE/CVF conference on computer vision and pattern recognition},
  pages={19967--19977},
  year={2023}
}

@inproceedings{decouple,
  title={Decoupling representation and classifier for long-tailed recognition},
  author={Kang, Bingyi and Xie, Saining and Rohrbach, Marcus and Yan, Zhicheng
          and Gordo, Albert and Feng, Jiashi and Kalantidis, Yannis},
  booktitle={Eighth International Conference on Learning Representations (ICLR)},
  year={2020}
}

@inproceedings{bbn,
  title={Bbn: Bilateral-branch network with cumulative learning for long-tailed visual recognition},
  author={Zhou, Boyan and Cui, Quan and Wei, Xiu-Shen and Chen, Zhao-Min},
  booktitle={Proceedings of the IEEE/CVF conference on computer vision and pattern recognition},
  pages={9719--9728},
  year={2020}
}

@inproceedings{paco,
  title={Parametric contrastive learning},
  author={Cui, Jiequan and Zhong, Zhisheng and Liu, Shu and Yu, Bei and Jia, Jiaya},
  booktitle={Proceedings of the IEEE/CVF international conference on computer vision},
  pages={715--724},
  year={2021}
}

@inproceedings{bcl,
  title={Balanced contrastive learning for long-tailed visual recognition},
  author={Zhu, Jianggang and Wang, Zheng and Chen, Jingjing and Chen, Yi-Ping Phoebe and Jiang, Yu-Gang},
  booktitle={Proceedings of the IEEE/CVF Conference on Computer Vision and Pattern Recognition},
  pages={6908--6917},
  year={2022}
}

@article{b_softmax,
  title={Balanced meta-softmax for long-tailed visual recognition},
  author={Ren, Jiawei and Yu, Cunjun and Ma, Xiao and Zhao, Haiyu and Yi, Shuai and others},
  journal={Advances in neural information processing systems},
  volume={33},
  pages={4175--4186},
  year={2020}
}

@article{logit_adjust,
  title={Long-tail learning via logit adjustment},
  author={Menon, Aditya Krishna and Jayasumana, Sadeep and Rawat, Ankit Singh and Jain, Himanshu and Veit, Andreas and Kumar, Sanjiv},
  journal={arXiv preprint arXiv:2007.07314},
  year={2020}
}

@inproceedings{oltr,
  title={Large-scale long-tailed recognition in an open world},
  author={Liu, Ziwei and Miao, Zhongqi and Zhan, Xiaohang and Wang, Jiayun and Gong, Boqing and Yu, Stella X},
  booktitle={Proceedings of the IEEE/CVF conference on computer vision and pattern recognition},
  pages={2537--2546},
  year={2019}
}

@article{ldam,
  title={Learning imbalanced datasets with label-distribution-aware margin loss},
  author={Cao, Kaidi and Wei, Colin and Gaidon, Adrien and Arechiga, Nikos and Ma, Tengyu},
  journal={Advances in neural information processing systems},
  volume={32},
  year={2019}
}

@inproceedings{cbloss,
  title={Class-balanced loss based on effective number of samples},
  author={Cui, Yin and Jia, Menglin and Lin, Tsung-Yi and Song, Yang and Belongie, Serge},
  booktitle={Proceedings of the IEEE/CVF conference on computer vision and pattern recognition},
  pages={9268--9277},
  year={2019}
}

@inproceedings{inaturalist,
  title={The inaturalist species classification and detection dataset},
  author={Van Horn, Grant and Mac Aodha, Oisin and Song, Yang and Cui, Yin and Sun, Chen and Shepard, Alex and Adam, Hartwig and Perona, Pietro and Belongie, Serge},
  booktitle={Proceedings of the IEEE conference on computer vision and pattern recognition},
  pages={8769--8778},
  year={2018}
}

@inproceedings{ncl,
  title={Nested collaborative learning for long-tailed visual recognition},
  author={Li, Jun and Tan, Zichang and Wan, Jun and Lei, Zhen and Guo, Guodong},
  booktitle={Proceedings of the IEEE/CVF Conference on Computer Vision and Pattern Recognition},
  pages={6949--6958},
  year={2022}
}

@article{reslt,
  title={Reslt: Residual learning for long-tailed recognition},
  author={Cui, Jiequan and Liu, Shu and Tian, Zhuotao and Zhong, Zhisheng and Jia, Jiaya},
  journal={IEEE transactions on pattern analysis and machine intelligence},
  volume={45},
  number={3},
  pages={3695--3706},
  year={2022},
  publisher={IEEE}
}

@article{sade,
  title={Self-supervised aggregation of diverse experts for test-agnostic long-tailed recognition},
  author={Zhang, Yifan and Hooi, Bryan and Hong, Lanqing and Feng, Jiashi},
  journal={Advances in Neural Information Processing Systems},
  volume={35},
  pages={34077--34090},
  year={2022}
}

@inproceedings{ace,
  title={Ace: Ally complementary experts for solving long-tailed recognition in one-shot},
  author={Cai, Jiarui and Wang, Yizhou and Hwang, Jenq-Neng},
  booktitle={Proceedings of the IEEE/CVF international conference on computer vision},
  pages={112--121},
  year={2021}
}

@article{ride,
  title={Long-tailed recognition by routing diverse distribution-aware experts},
  author={Wang, Xudong and Lian, Long and Miao, Zhongqi and Liu, Ziwei and Yu, Stella X},
  journal={arXiv preprint arXiv:2010.01809},
  year={2020}
}

@article{imagenet,
Author = {Olga Russakovsky and Jia Deng and Hao Su and Jonathan Krause and Sanjeev Satheesh and Sean Ma and Zhiheng Huang and Andrej Karpathy and Aditya Khosla and Michael Bernstein and Alexander C. Berg and Li Fei-Fei},
Title = {{ImageNet Large Scale Visual Recognition Challenge}},
Year = {2015},
journal   = {International Journal of Computer Vision (IJCV)},
volume={115},
number={3},
pages={211-252}
}

@article{cifar,
  title={Learning multiple layers of features from tiny images},
  author={Krizhevsky, Alex and Hinton, Geoffrey and others},
  year={2009},
  publisher={Toronto, ON, Canada}
}

@inproceedings{resnet,
  title={Deep residual learning for image recognition},
  author={He, Kaiming and Zhang, Xiangyu and Ren, Shaoqing and Sun, Jian},
  booktitle={Proceedings of the IEEE conference on computer vision and pattern recognition},
  pages={770--778},
  year={2016}
}

@inproceedings{resnext,
  title={Aggregated residual transformations for deep neural networks},
  author={Xie, Saining and Girshick, Ross and Doll{\'a}r, Piotr and Tu, Zhuowen and He, Kaiming},
  booktitle={Proceedings of the IEEE conference on computer vision and pattern recognition},
  pages={1492--1500},
  year={2017}
}

@article{pytorch,
  title={Pytorch: An imperative style, high-performance deep learning library},
  author={Paszke, Adam and Gross, Sam and Massa, Francisco and Lerer, Adam and Bradbury, James and Chanan, Gregory and Killeen, Trevor and Lin, Zeming and Gimelshein, Natalia and Antiga, Luca and others},
  journal={Advances in neural information processing systems},
  volume={32},
  year={2019}
}

@article{fixmatch,
  title={Fixmatch: Simplifying semi-supervised learning with consistency and confidence},
  author={Sohn, Kihyuk and Berthelot, David and Carlini, Nicholas and Zhang, Zizhao and Zhang, Han and Raffel, Colin A and Cubuk, Ekin Dogus and Kurakin, Alexey and Li, Chun-Liang},
  journal={Advances in neural information processing systems},
  volume={33},
  pages={596--608},
  year={2020}
}

@article{proco,
  title={Probabilistic contrastive learning for long-tailed visual recognition},
  author={Du, Chaoqun and Wang, Yulin and Song, Shiji and Huang, Gao},
  journal={IEEE Transactions on Pattern Analysis and Machine Intelligence},
  volume={46},
  number={9},
  pages={5890--5904},
  year={2024},
  publisher={IEEE}
}

@inproceedings{randaug,
  title={Randaugment: Practical automated data augmentation with a reduced search space},
  author={Cubuk, Ekin D and Zoph, Barret and Shlens, Jonathon and Le, Quoc V},
  booktitle={Proceedings of the IEEE/CVF conference on computer vision and pattern recognition workshops},
  pages={702--703},
  year={2020}
}

@article{autoaugment,
  title={Autoaugment: Learning augmentation policies from data},
  author={Cubuk, Ekin D and Zoph, Barret and Mane, Dandelion and Vasudevan, Vijay and Le, Quoc V},
  journal={arXiv preprint arXiv:1805.09501},
  year={2018}
}

@article{ncl++,
  title={NCL++: Nested collaborative learning for long-tailed visual recognition},
  author={Tan, Zichang and Li, Jun and Du, Jinhao and Wan, Jun and Lei, Zhen and Guo, Guodong},
  journal={Pattern Recognition},
  volume={147},
  pages={110064},
  year={2024},
  publisher={Elsevier}
}

@inproceedings{ltrl,
  title={LTRL: Boosting Long-tail Recognition via Reflective Learning},
  author={Zhao, Qihao and Dai, Yalun and Lin, Shen and Hu, Wei and Zhang, Fan and Liu, Jun},
  booktitle={European Conference on Computer Vision},
  pages={1--18},
  year={2024},
  organization={Springer}
}

@article{ecl,
  title={Towards effective collaborative learning in long-tailed recognition},
  author={Xu, Zhengzhuo and Chai, Zenghao and Xu, Chengyin and Yuan, Chun and Yang, Haiqin},
  journal={IEEE Transactions on Multimedia},
  volume={26},
  pages={3754--3764},
  year={2023},
  publisher={IEEE}
}

@inproceedings{ban,
  title={Born again neural networks},
  author={Furlanello, Tommaso and Lipton, Zachary and Tschannen, Michael and Itti, Laurent and Anandkumar, Anima},
  booktitle={International conference on machine learning},
  pages={1607--1616},
  year={2018},
  organization={PMLR}
}

@inproceedings{byot,
  title={Be your own teacher: Improve the performance of convolutional neural networks via self distillation},
  author={Zhang, Linfeng and Song, Jiebo and Gao, Anni and Chen, Jingwei and Bao, Chenglong and Ma, Kaisheng},
  booktitle={Proceedings of the IEEE/CVF international conference on computer vision},
  pages={3713--3722},
  year={2019}
}

@article{meanteacher,
  title={Mean teachers are better role models: Weight-averaged consistency targets improve semi-supervised deep learning results},
  author={Tarvainen, Antti and Valpola, Harri},
  journal={Advances in neural information processing systems},
  volume={30},
  year={2017}
}

@article{augmix,
  title={Augmix: A simple data processing method to improve robustness and uncertainty},
  author={Hendrycks, Dan and Mu, Norman and Cubuk, Ekin D and Zoph, Barret and Gilmer, Justin and Lakshminarayanan, Balaji},
  journal={arXiv preprint arXiv:1912.02781},
  year={2019}
}

@article{prl,
  title={Breaking Long-Tailed Learning Bottlenecks: A Controllable Paradigm with Hypernetwork-Generated Diverse Experts},
  author={Zhao, Zhe and Wen, HaiBin and Wang, Zikang and Wang, Pengkun and Wang, Fanfu and Lai, Song and Zhang, Qingfu and Wang, Yang},
  journal={Advances in Neural Information Processing Systems},
  volume={37},
  pages={7493--7520},
  year={2024}
}

@inproceedings{coco,
  title={Microsoft coco: Common objects in context},
  author={Lin, Tsung-Yi and Maire, Michael and Belongie, Serge and Hays, James and Perona, Pietro and Ramanan, Deva and Doll{\'a}r, Piotr and Zitnick, C Lawrence},
  booktitle={European conference on computer vision},
  pages={740--755},
  year={2014},
  organization={Springer}
}

@article{mixup,
  title={mixup: Beyond empirical risk minimization},
  author={Zhang, Hongyi and Cisse, Moustapha and Dauphin, Yann N and Lopez-Paz, David},
  journal={arXiv preprint arXiv:1710.09412},
  year={2017}
}

@article{medical,
  title={Towards long-tailed, multi-label disease classification from chest X-ray: Overview of the CXR-LT challenge},
  author={Holste, Gregory and Zhou, Yiliang and Wang, Song and Jaiswal, Ajay and Lin, Mingquan and Zhuge, Sherry and Yang, Yuzhe and Kim, Dongkyun and Nguyen-Mau, Trong-Hieu and Tran, Minh-Triet and others},
  journal={Medical Image Analysis},
  volume={97},
  pages={103224},
  year={2024},
  publisher={Elsevier}
}

@inproceedings{autonomous_driving,
  title={Towards long-tailed 3d detection},
  author={Peri, Neehar and Dave, Achal and Ramanan, Deva and Kong, Shu},
  booktitle={Conference on Robot Learning},
  pages={1904--1915},
  year={2023},
  organization={PMLR}
}

@inproceedings{cutmix,
  title={Cutmix: Regularization strategy to train strong classifiers with localizable features},
  author={Yun, Sangdoo and Han, Dongyoon and Oh, Seong Joon and Chun, Sanghyuk and Choe, Junsuk and Yoo, Youngjoon},
  booktitle={Proceedings of the IEEE/CVF international conference on computer vision},
  pages={6023--6032},
  year={2019}
}

@inproceedings{tsmof,
 author = {Zhao, Zhe and Gong, Zhiheng and Wang, Pengkun and Wen, HaiBin and Guo, Cankun and Xue, Bo and Lin, Xi and Wang, Zhenkun and Zhang, Qingfu and Wang, Yang},
 booktitle = {Advances in Neural Information Processing Systems},
 pages = {5224--5252},
 title = {TS-MOF: Two-Stage Multi-Objective Fine-tuning for Long-Tailed Recognition},
 volume = {38},
 year = {2025}
}

@inproceedings{icl,
  title={Enhancing mixture of experts with independent and collaborative learning for long-tail visual recognition},
  author={Chen, Yanhao and Jian, Zhongquan and Ke, Nianxin and Hu, Shuhao and Jiao, Junjie and Hong, Qingqi and Wu, Qingqiang},
  booktitle={Proceedings of the Thirty-Fourth International Joint Conference on Artificial Intelligence},
  pages={828--836},
  year={2025}
}

@inproceedings{glmc,
  title={Global and local mixture consistency cumulative learning for long-tailed visual recognitions},
  author={Du, Fei and Yang, Peng and Jia, Qi and Nan, Fengtao and Chen, Xiaoting and Yang, Yun},
  booktitle={Proceedings of the IEEE/CVF conference on computer vision and pattern recognition},
  pages={15814--15823},
  year={2023}
}

@inproceedings{ssl_expert,
  title={Three heads are better than one: Complementary experts for long-tailed semi-supervised learning},
  author={Ma, Chengcheng and Elezi, Ismail and Deng, Jiankang and Dong, Weiming and Xu, Changsheng},
  booktitle={Proceedings of the AAAI Conference on Artificial Intelligence},
  volume={38},
  number={13},
  pages={14229--14237},
  year={2024}
}

@inproceedings{mutual,
  title={Mutual learning for long-tailed recognition},
  author={Park, Changhwa and Yim, Junho and Jun, Eunji},
  booktitle={Proceedings of the IEEE/CVF winter conference on applications of computer vision},
  pages={2675--2684},
  year={2023}
}

@article{ensemble_diversity,
  title={A unified theory of diversity in ensemble learning},
  author={Wood, Danny and Mu, Tingting and Webb, Andrew M and Reeve, Henry WJ and Luj{\'a}n, Mikel and Brown, Gavin},
  journal={Journal of machine learning research},
  volume={24},
  number={359},
  pages={1--49},
  year={2023}
}

@inproceedings{shike,
  title={Long-tailed visual recognition via self-heterogeneous integration with knowledge excavation},
  author={Jin, Yan and Li, Mengke and Lu, Yang and Cheung, Yiu-ming and Wang, Hanzi},
  booktitle={Proceedings of the IEEE/CVF conference on computer vision and pattern recognition},
  pages={23695--23704},
  year={2023}
}

@inproceedings{pred_consistency,
author    = {Nan Kang and Hong Chang and Bingpeng MA and Shutao Bai and Shiguang Shan and Xilin Chen},
title     = {Predictive Consistency Learning for Long-Tailed Recognition},
booktitle = {34th British Machine Vision Conference 2023, {BMVC} 2023, Aberdeen, UK, November 20-24, 2023},
publisher = {BMVA},
year      = {2023},
}

@article{fourier,
  title={A fourier perspective on model robustness in computer vision},
  author={Yin, Dong and Gontijo Lopes, Raphael and Shlens, Jon and Cubuk, Ekin Dogus and Gilmer, Justin},
  journal={Advances in Neural Information Processing Systems},
  volume={32},
  year={2019}
}

@inproceedings{gradcam,
  title={Grad-cam: Visual explanations from deep networks via gradient-based localization},
  author={Selvaraju, Ramprasaath R and Cogswell, Michael and Das, Abhishek and Vedantam, Ramakrishna and Parikh, Devi and Batra, Dhruv},
  booktitle={Proceedings of the IEEE international conference on computer vision},
  pages={618--626},
  year={2017}
}

\end{document}